%% file: main.tex
\documentclass[runningheads]{llncs}
\usepackage[T1]{fontenc}
\usepackage{appendix}
\usepackage{graphicx}
\usepackage{amsmath}
\usepackage{subfigure}
\usepackage{array}
\usepackage{rotating}
\usepackage{multirow} 
\usepackage{wrapfig}
\usepackage{booktabs}
\usepackage{amsmath,amssymb}

\usepackage[table, usenames, dvipsnames]{xcolor}
\definecolor{dustygreen}{rgb}{0.60, 0.74, 0.62}
\definecolor{dustypink}{rgb}{0.87, 0.67, 0.70}
\definecolor{darkerdustygreen}{rgb}{0.45, 0.65, 0.50}
\definecolor{darkerdustypink}{rgb}{0.6, 0.35, 0.38}

\DeclareMathOperator{\EX}{\mathbb{E}}
\begin{document}
\title{Conceptual Contrastive Edits in Textual and Vision-Language Retrieval}
%\title{Explainable Hallucination Detection in Image Captioning through Conceptual Edits}

\author{Maria Lymperaiou\orcidID{0000-0001-9442-4186} \and Giorgos Stamou\orcidID{0000-0003-1210-9874}}

\authorrunning{M. Lymperaiou and G. Stamou}

\institute{National Technical University of Athens \\
\email{marialymp@ails.ece.ntua.gr} 
\email{gstam@cs.ntua.gr}}
\maketitle              
\begin{abstract}
As deep learning models grow in complexity, achieving model-agnostic interpretability becomes increasingly vital. In this work, we employ post-hoc conceptual contrastive edits to expose noteworthy patterns and biases imprinted in representations of retrieval models. We systematically design optimal and controllable contrastive interventions targeting various parts of speech, and effectively apply them to explain both linguistic and visiolinguistic pre-trained models in a black-box manner. Additionally, we introduce a novel metric to assess the per-word impact of contrastive interventions on model outcomes, providing a comprehensive evaluation of each intervention's effectiveness.

\end{abstract}

\section{Introduction}
In recent years, interpretability and explainability of linguistic and multimodal systems has become a persistent challenge and an unavoidable need \cite{mohammadi2025explainabilitypracticesurveyexplainable,dang2024explainableinterpretablemultimodallarge}. As models become larger and more opaque, black-box methods arise as the only viable solution for interpreting model inner workings and responses, since delving into their internals often becomes impossible \cite{gpt4,claude}. Even in the case of smaller pre-trained models, white-box explainability methods are deemed impractical, restricted to the specific-model's architecture and even impossible in cases that model weights are inaccessible, while performance may be influenced as well \cite{nlp-interpret-survey}. 

Such restriction inspire us to steer our efforts towards black-box explainability venues, pursuing model-agnostic, task-independent and adaptable solutions, enforced in a post-hoc manner upon already trained models. A simple but widely applicable idea incorporates black-box input-output relationship monitoring, which reveal explainable pathways imprinted during model pre-training. The impact of each sample can be better evaluated by applying input perturbations that indicate the outcome change as a function of the degree of perturbation. Such implementations are studied as a part of counterfactual and contrastive explanations \cite{Guidotti2024,contrastive-survey}. 
While input perturbations can be performed in many ways, from interfering with input features \cite{lime} to manipulating high-level concepts \cite{poeta2023conceptbasedexplainableartificialintelligence}, their resulting understandability varies when regarding human as the final recipient. For example, counterfactuals should be grounded to human-comprehensible semantics, otherwise equating themselves to adversarial examples \cite{browne2020semanticsexplanationcounterfactualexplanations}. At the same time, non-conceptual feature-based perturbations are proven fragile in the first place, resulting in untrustworthy and non-meaningful explanations \cite{ghorbani2018interpretationneuralnetworksfragile,unreliability}.

In our work, we focus on post-hoc, black-box conceptual explanations, applicable in a contrastive manner: implemented interventions contradict the default meaning of intervened samples, leading to semantically discernible changes. Contrastive explanations are considered to be more human-grounded \cite{MILLER20191}, amplifying semantic-related guarantees and holistic viewpoints offered by concepts \cite{poeta2023conceptbasedexplainableartificialintelligence}.

Getting back to the linguistic and multimodal domains, we discover that there is a severe lack of literature regarding conceptual contrastive explanations, especially when black-box and post-hoc desiderata are additionally enforced. Few endeavors address contrastive explanations in language generation \cite{jacovi-etal-2021-contrastive,yin-neubig-2022-interpreting}, revolving around the question \textit{'why did the model predict X instead of Y?'}. Moreover, contrastive edits on input text succeed in explaining text classification, even though the editing process is uncontrollable and not necessarily guaranteed \cite{ross-etal-2021-explaining,treviso-etal-2023-crest,polyjuice,ross2022tailorgeneratingperturbingtext,jiang-etal-2023-disco},  due to their reliance on large language models (LLMs), which are unpredictable on their own. 
Conversely, editors leveraging concept-driven edits \cite{lymperopoulos-etal-2024-optimal} can effortlessly provide controllability and optimality of edits, even if those may be compromised for efficiency.
Overall, existing editing frameworks, even task-agnostic ones, have been evaluated on classification settings exclusively \cite{morris2020textattackframeworkadversarialattacks,textfooler,ross-etal-2021-explaining,polyjuice,ross2022tailorgeneratingperturbingtext,jiang-etal-2023-disco}, questioning their extendability to non-classification tasks, even though substantial training procedures have been employed in all cases.

On the contrary, we deviate from classification tasks, targeting the widely underexplored field of semantic similarity, primarily experimenting on \textit{textual retrieval} \cite{lymperaiou-etal-2022-towards} using concept-driven contrastive edits. Then, we effortlessly extend our technique on \textit{text-to-image retrieval}, exhibiting the plug-and-play, zero-training and model-agnostic nature of our approach. Ultimately, we propose a highly efficient, adaptable, controllable and optimal post-hoc explanation framework for unimodal (textual) and vision-language (VL) retrieval models, incorporating several novelties outlined in the following sections.

\section{Background}
\subsection{Counterfactual and contrastive edits}
Counterfactual and contrastive edits for explainability lie close together since they scrutinize alternative outcomes elucidating model decision-making. In the interest of textual edits, prior literature indeed makes no clear distinction, circumventing formal definitions to focus on technical aspects instead. In fact, adversarial generations often fulfill the criteria for counterfactual/contrastive edits, leading them to be categorized under the same umbrella term.

In practice, granularity of edits ranges from word level \cite{textfooler,lymperopoulos-etal-2024-optimal} to sentence level \cite{sentence-level} perturbations, while more fine-grained levels (e.g. character-level edits \cite{hotflip}) naturally correspond to adversarial attacks, lacking semantic foundations. Attention-based feature importance measures to guide edits have been proposed in the recent work of Bhan et al \cite{Bhan2023TIGTEC}.
Nevertheless, most recent works employ masked language modeling to enable textual editing; for example, MiCE \cite{ross-etal-2021-explaining} performs a computationally expensive approach which however is heavily optimized towards achieving label-flipping of a pre-trained classifier via generating edits. Similar editors, such as CREST \cite{treviso-etal-2023-crest}, still require training to achieve appropriate editing. General-purpose editors deviate from optimization over pre-trained classifiers \cite{polyjuice,ross2022tailorgeneratingperturbingtext}, without eliminating though the significant computational overhead imposed by LLM fine-tuning. Prompting methods naturally emerge as a more viable LLM-related solution, inspiring the development of related textual editors for explainability \cite{dixit-etal-2022-core,jiang-etal-2023-disco,Bhattacharjee2023TowardsLC}. 
In any case, incorporating LLMs imposes variability over generated edits \cite{nguyen-etal-2024-ceval-benchmark}, introducing uncertainty to the outcome and therefore to the derived explanations overall.
A different direction suggests targeted edits directed via graph matching \cite{lymperopoulos-etal-2024-optimal}, enforcing advanced controllability and optimality. The same desiderata are enforced on edits addressing semantic similarity explanations \cite{lymperaiou-etal-2022-towards}, diverging from classification tasks for the first time.

Concept-based edits studied in this paper can be \textit{only} satisfied in the latter case, enabling word-level interventions which satisfy \textit{controllability} and \textit{optimality} constraints \cite{lymperaiou-etal-2022-towards,lymperopoulos-etal-2024-optimal}. Specifically, \textit{controllability} refers to the absolute requirement to substitute any concept, should this substitution is feasible; \textit{optimality} concerns providing the best solution obeying a certain notion of conceptual distance.
Such constraints are partially relaxed in \cite{lymperopoulos-etal-2024-optimal} as a sacrifice towards advanced efficiency: employing Graph Neural Networks (GNNs) \cite{gnn-survey} for concept assignment induces uncertainty and limits \textit{explainability} of resulting edits, referring to the traceability and meaningfulness behind each respective edit. In our current work, we resume the strict constraints to enforce \textit{explainability}, \textit{optimality} and \textit{controllability} in the proposed pipeline. This is a key novelty of our work, since those constrains have been previously imposed \textit{only} in \cite{lymperaiou-etal-2022-towards}, even though their performed interventions are simple and focused on certain adjectives. We expand the main idea to incorporate several parts of speech (POS) with the ultimate goal of explaining the impact of each POS on textual and VL retrieval.

\subsection{Combinatorial optimization on graphs}
\label{sec:combinatorial}
The pursue of \textit{optimality} for contrastive edits is an algorithmically exhaustive venture, calling for combinatorial optimization solutions \cite{pso} to eradicate uncertainty in the greatest level possible. This means that among an exponential solution space with respect to the number of participating concepts \cite{lymperopoulos-etal-2024-optimal}, optimization methods are necessitated in order to drop computational complexity to manageable levels. Consequently, we translate the problem of defining conceptual contrastive edits to finding the optimal match of concepts on a graph.

The heart of our framework considers a bipartite graph $\mathcal{G}=(V,E)$ whose vertices V can be divided in two disjoint sets $S$, $T$, while its edge set $E$ contains weighted edges of positive weights $w_{s\rightarrow t}\in \mathcal{W}$. In our case, we consider that $|S|\leq |T|$, enforcing all source concepts $s\in S$ to be matched, and thus substituted with some $t\in T$, if their in-between edge exists.
The \textbf{\textit{minimum weight bipartite matching}} instructs finding a matching $\mathcal{M}\subseteq E$ that pairs every node in $S$ with a unique node in $T$ such that the total weight of the matched edges $\sum^{\mathcal{M}}w_{s\rightarrow t}$ is minimized. Formally, the optimization problem can be expressed as:
\begin{equation*}
\begin{aligned}
    & \text{Minimize} \!&&\sum_{s\rightarrow t\in E} w_{s\rightarrow t} x_{s\rightarrow t},  
\quad x_{s\rightarrow t} =
\begin{cases}
    1 & \text{if edge } s\rightarrow t \text{ is included in $\mathcal{M}$}, \\
    0 & \text{otherwise}.
\end{cases}
\\
    & \text{subject to} \quad && \sum_{t \in T} x_{s\rightarrow t} = 1 \quad \forall s \in S, \qquad \qquad \qquad (1)\\
    & && \sum_{s \in S} x_{s\rightarrow t} \leq 1 \quad \forall t \in T, \qquad \qquad \qquad (2)\\
    %& && x_{s\rightarrow t} \in \{0, 1\} \quad \forall (s\rightarrow t) \in E \qquad \quad \:\;(3)
\end{aligned}
\end{equation*}
Solving the minimization function ensures \textit{optimality} of substitutions by considering the lightest possible $s\rightarrow t$ edges overall; $x_{s\rightarrow t}$ serves as an indicator variable that defines if the corresponding $s\rightarrow t$ edge participates in $\mathcal{M}$.
Constraint (1) imposes that each node $s\in S$ should be matched with some $t\in T$, enforcing \textit{controllability} of $s\rightarrow t$ substitutions. On the contrary, constraint (2) denotes that not all nodes $t\in T$ are going to be matched, since more than one substitution options may be available.

The minimum weight matching $\mathcal{M}$ can be found using the Hungarian algorithm \cite{hungarian_algo,min_weight_matching} with complexity $O(|S||T| \log |S|)$, ensuring execution \textit{efficiency}. The matching $\mathcal{M}$ is by nature \textit{explainable}, since all edges participating in $\mathcal{M}$ can be traced down due to the determinism of the Hungarian algorithm; therefore, this bipartite structural choice leads to \textit{explainable} concept substitutions, satisfying one of the key requirements of our proposed contrastive edits method.

\section{Method overview}
\begin{figure}[h!]
    \centering
    \includegraphics[width=\linewidth]{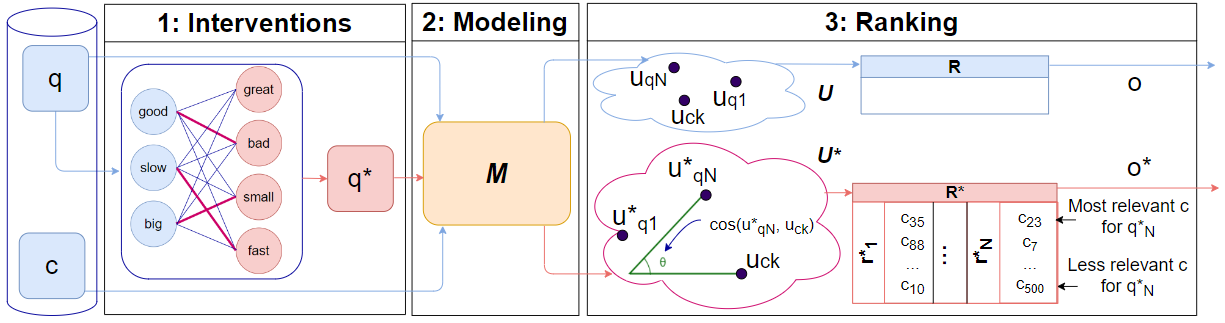}
    \caption{The pipeline of our contrastive edits method for retrieval models. Red is for the \textit{contrastively edited} stream, blue is for the \textit{default} stream.}
    \label{fig:flow}
    \vskip -0.2cm
\end{figure}

The workflow of our method, targeting either textual or VL retrieval (Fig. \ref{fig:flow}) comprises three stages. A dataset $D$ containing textual queries $q$ and corpora $c$ (either textual or visual for each application case) acts as the input of the workflow. In the first stage, a set of \textit{linguistic interventions} is applied on specific words to produce contrastive queries $q^*$. Specifically, words are extracted automatically from $q$ based on their part of speech (POS), or any other predefined characteristic, and are substituted with their counterparts (matched concepts on the bipartite graph $\mathcal{G}$), while the rest of the initial sentence remains unchanged.

The second stage involves the representation of $q^*$ and $c$ in the embedding space $U^*$ via a pre-trained model $M$. Language models are employed in the unimodal case (when $c$ is purely textual) and VL transformers are utilized in the multimodal case (when $c$ refers to images). 

Cosine similarity scores between all $u_{q^*}, u_c \in U^*$ embedding pairs for $q^*$ and $c$ respectively are obtained in the third stage, providing a rank $r^*_{q^*}$ of all corpora per $q^*$. The alternative outcome $o^*$ for all $q^*$ is measured using any ranking metric.  In the same sense, we drop the $^*$ notation to outline the default stream (without interventions) of our method.
\subsection{Non-minimal interventions for contrastive edits}
In order to stress the behavior of retrieval models $M$, semantically \textit{non-minimal} word-level interventions need to be applied on input $q$, resulting in $q^*$ with contrastive content. For example, replacing an adjective with its antonym satisfies such requirements: the non-minimal intervention is achieved by substituting with the most dissimilar word that preserves the existing POS. At this point, any observed alternative outcome $o^*$ compared to the default outcome $o$ is attributed to this antonym substitution. To implement non-minimal interventions, we consider three distinct cases, analyzed in the following paragraphs.

\paragraph{Graph-based edits}
In this case, the edits are performed using the minimum weight bipartite matching described in Section \ref{sec:combinatorial}.
Any word belonging to a specific POS at a time is extracted from $q$ and placed on the source set $S$ of the bipartite $\mathcal{G}$, while a copy of $S$ concepts comprises the target set $T$.
%or concepts provided by external knowledge sources, such as WordNet \cite{miller1995wordnet}. 
Interventions are realized by $s \rightarrow t$ concept substitutions, with feasible substitutions denoted by the existence of an edge $s \rightarrow t \in E$. Initially, all possible $s \rightarrow t$ paths are considered, resulting in a dense bipartite graph $\mathcal{G}$. Edge weights $w_{s\rightarrow t}\in W$ express a measure of similarity between $S$ and $T$ concepts; this similarity can be provided by external knowledge sources, such as WordNet \cite{miller1995wordnet}, or color-relatedness hierarchies \cite{lymperaiou-etal-2022-towards},
where conceptual distances are well-defined and deterministic.
In the case of WordNet, path similarity function provides a value ranging between 0 and 1 based on the shortest path that connects two given concepts within the Wordnet hierarchy. A \textit{lower} path similarity value between two concepts indicates higher conceptual \textit{dissimilarity}-which is our desired case for non-minimal interventions, resolved via \textit{minimum weight bipartite matching}. 

Apart from instructing specific optimizations, knowledge graph-driven edits that directly traverse the given knowledge graph are employed. This way, we acquire semantic relationships by exploiting the existing hierarchy which provides us with hypernym (more generic concepts), hyponym (more specific concepts) or antonym relationships. 
\paragraph{Permutations and swaps} consider words belonging in the same POS and target to shuffle their ordering to test the impact of positional variability. In fact, permutations constitute \textit{randomized matchings}, since there is no criterion on how to substitute a concept with another.
\paragraph{Concept Deletions} denote the removal of related semantics from a description, again targeting a specific POS at a time. This serves as a baseline to evaluate whether the existence of specific words is influential for the overall meaning, or if the model can infer missing information based on context.

\subsection{Implementing POS-specific interventions}
\label{sec:interventions}
Our interventions target independently adjectives (ADJ), nouns (NOUN), verbs (VERB) and adpositions (ADP) of $q$, aiming to produce as semantically distant as possible contrastively edited $q^*$. This requirement pushes $o^*$ to diverge from $o$, so that the influence of an intervention becomes evident during evaluation. 

First of all, we craft a variety of \textit{graph-based edits} exploiting our optimization procedure via bipartite matching. Regarding adjectives (ADJ), we replicate the \textit{\textbf{color-all (CA), color-in (CI), sizes (S), antonyms (A)}} experiments of \cite{lymperaiou-etal-2022-towards}. \textit{\textbf{CA}} and \textit{\textbf{CI}} replace any color in $q$ with the most semantically distant color from Matplotlib\footnote{https://matplotlib.org/3.3.3/gallery/color/named\_colors.html} or $D$ respectively. 
\textit{\textbf{Sizes (S)}} substitutes any of \{\textit{large, big, enormous, huge}\} with any of \{\textit{small, little, minor, tiny}\} and vice versa. \textit{\textbf{Antonyms (A)}} intervention (applicable to ADJ and VERB) substitutes concepts with their antonyms via WordNet. 

\textit{\textbf{External (E)}} substitutions (applicable to ADJ, NOUN, VERB) pair relevant concepts with their most distant matches in $D$ according to WordNet, obtaining the optimal matching $\mathcal{M}$ for each POS. Verbs contain multiple POS tags corresponding to particular grammatical cases, including verb tense, past participle, gerund etc. In that sense, \textbf{\textit{VERB-E}} intervention  ignores POS tags by default and places all verbs under the same concept graph. We also provide a combined variant named \textit{\textbf{VERB-E-comb}} which independently matches verbs per tag, forming separate concepts graphs, and then merges proposed substitutions to form $q^*$. Regarding adpositions (ADP) \textit{random} substitutions are sufficient, as there is no standard measure of their similarity.

By directly traversing WordNet, \textit{\textbf{hypernym (HE)}} and \textit{\textbf{hyponym (HO)}} substitutions (applicable to NOUN, VERB) regard more generic or more specific concept substitutions respectively, based on existing hierarchical relationships. 

Regarding \textit{permutations}, we focus on one POS at a time, crafting \textit{\textbf{ADJ-P}}, \textit{\textbf{NOUN-P}}, \textit{\textbf{VERB-P}} interventions; in the case of nouns, \textit{\textbf{NOUN-P}} preserves singulars and plurals, instructing substitutions that exclusively consider one grammatical number at a time; singular and plural substitutions are then merged to form $q^*$. Alternatively, we experiment with mixing grammatical numbers, implementing a variant that randomly permutes nouns, named  \textit{\textbf{NOUN-RP}} regardless of their grammatical number. Additionally, explicitly swapping singulars with plurals and vice versa (\textit{\textbf{singular plural swap-SPS}} experiment) is attempted to check model response to grammatical number perturbations. 

\textit{Concept deletion} targets all four POS (ADJ, NOUN, VERB, ADP); relevant experiments contain the characterization code \textit{\textbf{blank (B)}}. For example, \textbf{\textit{NOUN-B}} regards the deletion of all nouns within a sentence.

As special use cases, experiments bearing the \textit{\textbf{-sing}} code are only applied on singulars, e.g. \textit{\textbf{NOUN external sing}}. As an example, \textit{\textbf{NOUN E-sing}} implements optimal NOUN substitutions in singular number. 

In all aforementioned cases, we intervene on \textit{all} semantics belonging in the respective POS. For example, if a sentence contains five nouns, an intervention bearing the \textbf{\textit{NOUN}} code will be applied on all five nouns. This approach maximizes the contrastive nature of our performed interventions. Nevertheless, we consider altering \textit{only one} related semantic at a time to compare the effect between the minimum and maximum number of related edits.
In these cases, the characterization code \textit{\textbf{single (SG)}} is prepended to the experiment name. For instance, \textbf{NOUN single external (NOUN SG-E)} intervenes on one noun from the sentence, substituting it with its most semantically distant pairing.

We present some example regarding our contrastive edits in Appendix \ref{sec:contrastive-examples}.

\subsection{Contrastive edits may lead to invariant ranking outcomes}
An interesting observation regarding non-minimal interventions in ranking tasks is that the final metrics remain \textit{almost invariant} to such contrastive semantic edits \cite{lymperaiou-etal-2022-towards}. This is a surprising piece of evidence since such contrastive edits often result in non-meaningful queries, e.g. \textit{'the \underline{soft} rock fell from the cliff'}, where the contrastive edit $\textit{hard}\rightarrow \textit{soft}$ led to a rather contradicting sentence according to common sense. At this point, the value of our proposed explainability method arises: \textit{contrastive edits that result in invariant $o^*$ \textit{vs} $o$ outcomes reveal concepts that are overtaken from the ranking model, since non-minimal changes do not significantly influence its decision in many cases}. 

This invariability suggests that $M$ focuses on a limited  (and even questionable) concept set for retrieval, which may differ from those that humans deem as important. For example, an antonym substitution may totally inverse the meaning of a sentence from the human's perspective, while the same does not apply to the retrieval model $M$. Moreover, invariance indicates that $M$ is overly rigid or insensitive to variations in the input data. This insensitivity can be problematic, especially since the performed contrastive edits often introduce  changes that should logically affect the ranking. In total, \textit{the contrastive edit-outcome invariance} pattern enhances post-hoc explainability of models tested, especially since the editing procedure is applied on different POS at a time, targeting to unveil \textit{which POS are more influential to model decision-making}.
%A way to probe word importance without any access to the model is to contrastively intervene over concepts of a certain POS at a time, substituting or deleting relevant words. Achieving the lightest matching of $\textit{min}(W^M_n)$ remains the ultimate goal for all optimal interventions. Then, the same downstream task should be repeated after each intervention for all models \textit{M}. 
%In the case of ranking, we gradually perturb the number of concepts $n_i$ to be intervened per $q_i$, while enforcing $\textit{min}(W_n)$ for all interventions. 
This process reveals the effect of each intervention on the final outcome of the ranking stage, with interventions over more important -from the models' point of view- POS causing more intense outcome changes, as measured by the degree of change between $o$ and $o^*$.

\subsection{Measuring the effect of contrastive edits}
POS-related interventions result in a different number of per intervention perturbed words $n$; for example, the number of verbs may greatly differ from the number of nouns present in sentences of a dataset (Appendix \ref{sec:num_words_POS}), therefore verb-related interventions yield a disproportional number of source concepts $n$ to be substituted compared to noun-related interventions. This issue obfuscates the reason behind an alternative outcome $o^*\neq o$: is the influence of the intervention itself or is it just the number of words changed? To counter this imbalance, we propose a metric of \textit{Average Contrastive Effect (ACE)} that measures the per word influence of each intervention on the \textit{outcome change} $o \rightarrow o^*$. Mathematically, \textit{ACE} is expressed as the ratio between the average percentage of the \textit{outcome change} over the actual total number \textit{n} of words perturbed in all $q$:
\begin{equation}
  \textit{ACE}=\frac{\EX[|o - o^*|/o]}{n}\times scale   
\end{equation}

Intuitively, the more influential a POS is with respect to the selected model $M$, the higher its \textit{ACE} scores will be.
\textit{ACE} can be adjusted to any metric used to evaluate a task. The \textit{scale} factor can be any power of 10 (consistent across all experiments) and is used for visualization reasons, given that \textit{CE} scores may be small enough, when considering the large number of perturbed concepts $n$.

\section{Experiments}
We focus on the following types of retrieval: linguistic retrieval (\textbf{LR}), where $c$ is purely textual, and VL retrieval where $c$ is an image. Specifically, in text-image retrieval (\textbf{TIR}), a linguistic query retrieves an image, and inversely in image-text retrieval (\textbf{ITR}), an image retrieves the corresponding text. We focus on visual vocabularies as the dataset $D$ by utilizing the Flickr dataset \cite{flickr}, which comprises images accompanied by textual captions. Under this experimentation, we present the influence of our framework over VL models, while maintaining a fair comparison with fully linguistic ones. In all \textbf{LR}, \textbf{TIR}, \textbf{ITR} experiments the first sentence out of the 5 captions per image in $D$ serves as the query $q$, upon which all interventions are applied. The rest 4 sentences per image are used as the textual corpus $c$ for \textbf{LR} and remains unchanged. In \textbf{TIR}/\textbf{ITR} experiments, the ground truth Flickr  images are used as the corpus $c$.

\subsection{Models}
In the VL setting (\textbf{TIR}/\textbf{ITR}), CLIP \cite{clip} serves as the model $M$. Regarding \textbf{LR}, any semantic similarity model (as in \cite{lymperaiou-etal-2022-towards}) can be used. We provide results on various SBERT models \cite{sbert} (fine-tuned for semantic similarity). In all cases, we perform no training or fine-tuning on $M$, and we regard it as a black box.
\subsection{Evaluation}
The outcome of the ranking stage is evaluated using \textit{Recall@k (R@k)} reformed within our proposed \textit{ACE} metric, resulting in the $ACE_{R@k}$ metric, with $o$\textit{=R@k} for the ground truth rank $R$, $o^*$\textit{=}$R^*$\textit{@k} for the alternative rank $R^*$ and $n$ as the number of words substituted per intervention.  Finally, we select  scale=$10^5$ for \textit{ACE}; this is a random selection, consistently applied in all experiments to eliminate the presence of unecessarily large numbers of decimal digits.

\subsection{Text retrieval results}
Results regarding text retrieval using unimodal retrievers are presented per POS in separate Tables due to page limits. Specifically, Tables \ref{tab:adj-table}, \ref{tab:adj-table-sing} present results on adjectives (\textbf{\textit{ADJ}}), considering both singular and plural numbers in the first case, and singulars exclusively in the latter. Moreover, Table \ref{tab:adposition} refers to adpositions (\textbf{\textit{ADP}}), while Tables \ref{tab:noun1}, \ref{tab:noun2} refer to nouns (\textbf{\textit{NOUN}}), and Tables \ref{tab:verb1}, \ref{tab:verb2} to verbs (\textbf{\textit{VERB}}). Based on the value distributions of ACE scores, we highlight with \colorbox{dustygreen}{green} cells with high ACE scores ($ACE_{R@1}\geq 4$), and reversely with \colorbox{dustypink}{pink} cells with low ACE scores ($ACE_{R@1}<1$), drawing the attention on more or less influential interventions respectively. 
\input{adj_table}
\input{adj_table_sing}
\input{adposition}
\input{noun1}
\input{noun2}
\input{verb1}
\input{verb2}
\subsubsection{I. Analysis per POS}

The colored patterns in all presented Tables easily reveal some interesting POS-related insights. First of all, adjective interventions (Tables \ref{tab:adj-table}, \ref{tab:adj-table-sing}) outline some sparse findings, since varying behaviors emerge per model: for example, antonymic substitutions, including \textbf{\textit{Antonym (A)}} in Table \ref{tab:adj-table} and \textbf{\textit{Single Antonym (SG-A)}} in Table \ref{tab:adj-table-sing} achieve \colorbox{dustygreen}{high} ACE scores in some cases, but not in the majority of them. On the other hand, \textbf{\textit{size (S)}} substitutions present \colorbox{dustypink}{low} ACE scores for some models, but again not leading to a universal pattern.
A notable exception arises for permutations (\textbf{\textit{P}} column in Table \ref{tab:adj-table}), which achieve a low ACE score for all models, as indicated in \colorbox{dustypink}{highlighted} cells.

A striking pattern is revealed when focusing on adpositions on Table \ref{tab:adposition}: all interventions for all models score very \colorbox{dustypink}{low} in ACE metric, denoting the negligible influence of intervening over this specific POS. This finding is perfectly aligned with the VL case, since adposition interventions are extremely weak when delving into CLIP ranking results in Figure \ref{fig:clip-ace}, confirming that they are not crucial components in a visual description, regardless the query or candidate modality.

Nouns also present clear patterns more often than not: permutational interventions, such as \textbf{\textit{Permutations
(P)}}, \textbf{\textit{Sing.-plural (SPS)}} and \textbf{\textit{Random Permutations (RP)}} in Table \ref{tab:noun1} consistently illustrate very \colorbox{dustypink}{low} ACE scores. 
Conversely, \textbf{\textit{external singulars (E-sing)}} in Table \ref{tab:noun2} achieve \colorbox{dustygreen}{high} ACE results for most models, denoting that substituting singulars only is a highly influential intervention in comparison to intervening on nouns regardless of their grammatical number, as represented by the \textbf{\textit{external (E)}} intervention of Table \ref{tab:noun1}, where we observe the high-ACE effect to fade out. Therefore, plural substitutions are not as influential, exposing an asymmetrical bias regarding grammatical number for most models. 
Sparsely strong outcome changes are observed for some models regarding  \textbf{\textit{single external (SG-E)}} and \textbf{\textit{single
blank (SG-B)}} interventions, demonstrating that in a few cases intervening on \textit{a single noun} is analogously more influential that intervening on all nouns within the description.

In the case of verbs, findings are mixed, presenting the \colorbox{dustygreen}{highest} and the \colorbox{dustypink}{lowest} ACE scores for the same interventions, when different models are considered: for example, this is observed in verb \textbf{\textit{permutations (P)}}, \textbf{\textit{hypernyms (HE)}}, \textbf{\textit{hyponyms (HO)}} and \textbf{\textit{external (E)}} substitutions in Table \ref{tab:verb1}.
On the other hand, \textbf{\textit{external
combined (E-comb)}} interventions mostly exhibit \colorbox{dustygreen}{high} ACE scores, revealing that more intense outcome changes are attained when verb tags are respected during substitutions. The opposite behavior regarding interventional influence expressed via ACE is observed in Table \ref{tab:verb2},  \textbf{\textit{single external (SG-E)}} column, illustrating that substituting a \textit{single} verb at a time is not adequate to perceptibly alter ranking outcomes. Other than that, antonymic verb substitutions such as \textbf{\textit{antonym
(A)}} and \textbf{\textit{single antonym
(SG-A)}}, as well as verb deletions, such as \textbf{\textit{blank (B)}} and \textbf{\textit{single blank (SG-B)}} pose a mediocre or rather negligible outcome change in most cases.

\subsubsection{II. Analysis per intervention}
One of the interventional strategies posing more consistent results refers to permutations, leading to minor ranking outcome changes in adjectives (Table \ref{tab:adj-table} - \textbf{\textit{permutation (P)}} column) and  nouns (Table \ref{tab:noun1}, \textbf{\textit{permutations (P)}}
\textbf{\textit{singular-plural swaps (SPS)}} and \textbf{\textit{random permutations (RP)}}). This finding agrees with the results derived from CLIP in the multimodal retrieval case (Figure \ref{fig:clip-ace}), proving that word ordering invariance  affects textual retrieval for either NOUN, ADJ or VERB POS regardless of the participating modalities. 

Insensitivity to word ordering is rather unexpected based on common sense, as the meaning of a sentence can completely change when nouns are permuted. However, this \textbf{order-invariance} can be attributed to the way SBERT models are trained: Unlike autoregressive models, SBERT does not process words in a fixed left-to-right order. Instead, self-attention aggregates information from all words simultaneously, making it more order-invariant, especially for non-grammatical or minor word swaps, as the aforementioned noun-preserving permutations. Furthermore, SBERT generates sentence embeddings using mean pooling (or other pooling strategies). This process smooths out word order effects, meaning that two sentences with the same words but in different orders can end up with nearly identical embeddings. However, this finding denotes that current linguistic representations for semantic similarity are somewhat problematic, since meaningful and non-meaningful sentences lie close together in the embedding space. \colorbox{dustypink}{Low} ACE scores regarding \textbf{\textit{SPS}} interventions also reveal that SBERT models are \textbf{number-invariant} to some degree. This phenomenon can be possibly attributed to how tokenization is performed: for example, words like cat and cats share overlapping subword tokens (cat and \#\#s). The minor difference in representation leads to minimal impact on sentence embeddings. Other than that, SBERT focuses more on standalone semantic similarity rather than exact word forms. Since singular and plural forms often convey nearly the same meaning in most contexts, the model does not strongly differentiate between them. Notably, SBERT's over-reliance on pure semantic similarity rather than syntactic and grammatical structure means it might fail to distinguish sentences that have different meanings due to word order or number changes, once again indicating a possible direction for improvement regarding linguistic representations. Overall, our permutation experiments highlight the -unresolved so far- challenges of accurately representing human language in an embedding space. %Ideally, despite containing the same semantics, representations of fully correct versus semantically meaningless phrasings should be placed further apart within the embedding space they are currently projected. 
The permutational effect is less prominent for verbs, as denoted in \textbf{\textit{permutations (P)}} column of Table \ref{tab:verb1}, indicating an unexpected representation choice made from a few SBERT models which are surprisingly more sensitive in verb ordering, as in the case of stsb-roberta-base and nq-distilbert.

External substitutions (containing the code \textbf{\textit{E}}) arise as very influential interventions in noun or verb substitutions, but not for adjectives, where the effect is rather mediocre. For example, in Table \ref{tab:noun2}, external noun substitutions considering singulars exclusively (\textbf{\textit{E-sing}} column) comprise the highest-ACE intervention for the majority of semantic similarity models used, while a similar result holds for \textbf{\textit{external
combined (E-comb)}} verb substitutions in Table \ref{tab:verb1}. Those results are somehow expected since we enforce the most intense substitutions, ensuring non-minimal semantic changes, as well as intervening over all related semantics that refer to each specific POS. Under those observations, we confirm the importance of our proposed graph-based framework towards implementing contrastive edits which are ultimately able to reveal the importance different models pose on certain words.

\textbf{\textit{Hypernym (HE)}} or \textbf{\textit{hyponym (HO)}} interventions for nouns are neutral overall, suggesting that more intense substitutions are required to derail the retrieval process, something confirmed when implementing \textbf{\textit{external (E)}} manipulations.
Moreover, substituting verbs with their hyponyms, as denoted in the \textbf{\textit{VERB hyponym (HO)}} column of Table \ref{tab:verb1} is highly influential in most cases, underlining that verb overspecialization  is mostly adequate to severely break the semantic inter-relationships, even more than when substituting them with the most dissimilar verbs in WordNet, as indicated by the \colorbox{dustygreen}{higher} ACE scores of \textbf{\textit{HO}} interventions. The inverse direction of specialization i.e. substituting with more general concepts is almost negligible in comparison, as revealed by the lower ACE scores in the \textbf{\textit{Hypernyms (HE)}} column of Table \ref{tab:verb1}. 

Concept deletions also pose uninteresting patterns, almost always achieving intermediate ACE scores for either adjectives (\textbf{\textit{blank (B)}} column - Table \ref{tab:adj-table}, \textbf{\textit{single blank (SG-B)}} column - Table \ref{tab:adj-table-sing}), nouns (\textbf{\textit{blank (B)}}, \textbf{\textit{single blank (SG-B)}} - Table \ref{tab:noun1}) or verbs (\textbf{\textit{blank (B)}}, \textbf{\textit{single blank (SG-B)}} - Table \ref{tab:verb2}). However, ACE scores are \colorbox{dustygreen}{higher} for noun deletions in comparison to adjectives or verbs, exhibiting the also higher importance of removing verbs over the other POS. This is an expected pattern, as nouns are predominant when describing images, therefore their absence is harder to be inferred. 

Also in a technical level, SBERT models, being trained on sentence-level semantic similarity tasks, prioritize content words that define entities and objects (nouns) over modifiers (adjectives) or actions (verbs). From the perspective of the SBERT pre-training datasets, like NLI and STS,  similarity is judged at the sentence level. Many paraphrases in training involve modifications rather than complete shifts in meaning, making the model robust to slight deletions. Other than that, SBERT’s mean pooling mechanism over word embeddings smooths out the impact of individual words. This results in reduced sensitivity to functionally "weaker" words (adjectives, some verbs) and stronger impact from deleting core content words (nouns), but with context compensation. 

Overall, we can conclude that in most cases, the semantic similarity models are able to derive the meaning of the sentence exploiting conceptualization, therefore ignoring the missing words, even in cases when all related instances of a certain POS are eliminated. Even though ignoring deleted semantics can be associated with model robustness and flexibility, implications due to the closely lying representations can also occur. When precision is required, as in the case of retrieval in more critical fields, such as in medical or legal cases, information loss can be associated with severe consequences. For example, fine-grained distinctions may be lost, such as prioritizing the retrieval of phrasings "urgent surgery required" vs. "surgery required".
\subsection{VL retrieval Results}
\begin{figure}[h]
    \centering \vskip -0.1cm
\includegraphics[width=1.02\linewidth]{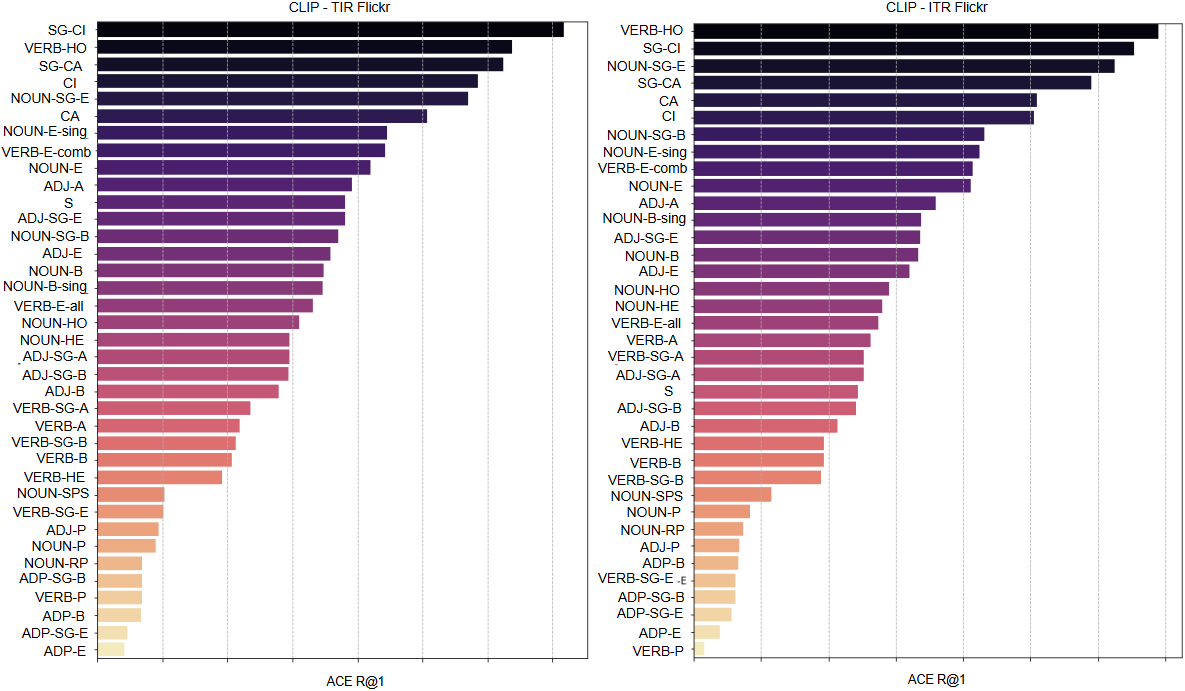}
    \caption{Results ($ACE_{R@1}$ metric) for text-image retrieval (\textbf{TIR}) on the left and image-text retrieval (\textbf{ITR}) on the right on Flickr dataset for all interventions.}
    \label{fig:clip-ace}
    \vskip -0.1cm
\end{figure}

We present ranking results on Flickr using the $ACE_{R@1}$ metric for all interventions in Figure \ref{fig:clip-ace} for \textbf{TIR} (left) and \textbf{ITR} (right). In all cases, higher ACE numbers
denote more influential interventions.

Interestingly, the two directions of retrieval differ in terms of ACE R@1 metrics driven from varying interventions. This observation means that despite image and text elements being mapped on the same semantic space,
the distribution of embeddings may be different for images versus texts. For example, image embeddings might be more clustered or have different variance than text embeddings, affecting similarity scores differently depending on which modality is used as the query.
Such differences in the similarity distributions may be due to  inherent properties of each modality, such as dimensionality, normalization, or noise characteristics. A metric such as cosine similarity might behave differently when comparing an image query to text candidates versus a text query to image candidates.

Nevertheless, it is apparent that similar interventions may rank higher or lower consistently in both retrieval scenarios. For instance, substituting verbs with their hyponyms (\textbf{\textit{VERB-HO}}) is the most influential intervention in \textbf{ITR} and the second most influential in \textbf{TIR}, meaning that when overspecifying verbs, VL retrieval is altered the most. At the same time, many adjective-related intervention also appear very predominant, such as \textbf{\textit{SG-CI}} and \textbf{\textit{SG-CA}} (substituting colors with other internal/external dataset colors respectively, but considering singulars only), as well as \textbf{\textit{CI}} and \textbf{\textit{CA}} (similar to the previous intervention, but the singular/plural number is not considered). We expect colors to be an influential attribute in visual vocabularies, since objects in scenes are characterized by distinct colors; therefore, breaking the object-color relationship poses a non-negligible semantic perturbation, expectedly leading to an observable alternative outcome change. This finding comes in contrast with \textbf{LR} results, where color-related inteventions appeared weaker, under ACE evaluation. This observation denotes a significant difference in handling visual information in textual versus VL scenarios: apparently, missing the visual modality leads to some loss of information which cannot be compensated by purely textual data.

On the other hand, many adpositional-related perturbations, such as \textbf{\textit{ADP-B}} and \textbf{\textit{ADP-E}} lie lower in ACE for both \textbf{TIR} and \textbf{ITR}, denoting the weak influence that this part of speech poses in ranking results. We can confirm that as in \textbf{LR}, adpositions are not able on their own to severely altering the semantic meaning of a sentence, at least in comparison to adjectives, nouns or verbs.

There are also some notable differences between \textbf{TIR} and \textbf{ITR}, as presented in the plots of Figure \ref{fig:clip-ace}. In \textbf{TIR}, noun-related elements (e.g., \textbf{\textit{NOUN-E-sing}},\textbf{\textit{ NOUN-E}}, \textbf{\textit{NOUN-SG-B}}) are relatively high-ranked, indicating that intervening in related words has a strong effect on retrieval.
In \textbf{ITR}, verb-related elements (e.g., \textbf{\textit{VERB-HO}},\textbf{\textit{ VERB-E-comb}}, \textbf{\textit{VERB-B}}, \textbf{\textit{VERB-SG-A}}) seem to have a slightly stronger influence compared to \textbf{TIR}. 
Based on these insights, 
\textbf{TIR} may rely more on noun-based grounding, meaning object descriptions are more crucial, while
\textbf{ITR} appears more sensitive to verbs and actions, possibly because descriptions of actions vary more depending on the image content. 

Low ACE influence in both \textbf{TIR} and \textbf{ITR} is observed in noun permutation interventions, such as \textbf{\textit{NOUN-P}} and \textbf{\textit{NOUN-RP}}, indicating the insensitivity of CLIP towards noun order within a visual description. This finding agrees with how SBERT models handle permutations in \textbf{LR} experiments. In these cases, the grammatical role of nouns stays unchanged, even if the specific nouns swap places. Since also VL retrieval models rely heavily on positional embeddings and sentence structure, this may lead to minimal disruption. Furthermore, minimally altered ranks reveal the excessive trust on semantics per-se rather than their positioning, even though the overall meaning of the sentence is broken after related interventions are applied. In that case, retrieval will still regard the related semantics correctly, even though their contextuals are disrupted. The distributed nature of contextual embeddings produced from CLIP may justify this invariance, since the global meaning of a description is not severely altered -at least from the model's perspective.
Verb and adjective permutations (\textbf{\textit{VERB-P}} and \textbf{\textit{ADJ-P }}respectively) also present low ACE scores, proving that invariance also concerns verb and adjective ordering. In many cases, model pre-training on datasets such as COCO \cite{coco} or Flickr \cite{flickr} exposes models to the same verb-noun and adjective-noun pairings; if a swap still maintains common co-occurrences, retrieval accuracy may not be strongly affected overall. This correlational nature appears to be more important than preserving the meaning of the description, indicating the model's anchoring on standalone concepts rather than their semantic interconnections.
Another low-influential intervention regarding verbs is \textbf{\textit{VERB-HE}}, where a verb is substituted with its hypernym; therefore, generalizing verb concepts is not significantly impactful, something that agrees with common sense. This finding mostly agrees with LR results for \textbf{\textit{VERB HE}}.

Overall, such differences in ranking behavior reinforce that retrieval asymmetry exists despite a shared embedding space for the two modalities. Other than that, many  interventional patterns remain consistent across purely textual and VL retrieval, with the notable exception of interventions concerning purely visual concepts, such as the ones involving colors.

\section{Conclusion}
In this work, we present a novel post-hoc explainability framework that harnesses knowledge graphs to craft contrastive conceptual edits. Our interventions pose the benefits of controllability, optimality, explainability and computational efficiency, applicable in a model-agnostic way.
We apply the proposed framework on multimodal and unimodal retrieval scenarios, outlining a large amount of patterns hidden within model representations. Overall, we conclude that in several cases, pre-trained retrieval models, widely used in research and practice, present over-reliance on certain semantics over others, while often ignoring word ordering when the semantics of interest are present. 

\begin{credits}
\subsubsection{\ackname} This work was supported by the Hellenic Foundation for Research and Innovation (HFRI) under the 3rd Call for HFRI PhD Fellowships (Fellowship Number 5537).

\subsubsection{\discintname}
The authors have no competing interests to declare that are
relevant to the content of this article. 
\end{credits}

\bibliographystyle{splncs04}
\bibliography{main}
\appendix
\section{Dataset details}\label{sec:num_words_POS}

\begin{figure}[t!]
    \centering
    % First subplot
    \subfigure[Per caption numbers.]{
        \includegraphics[width=0.4\textwidth]{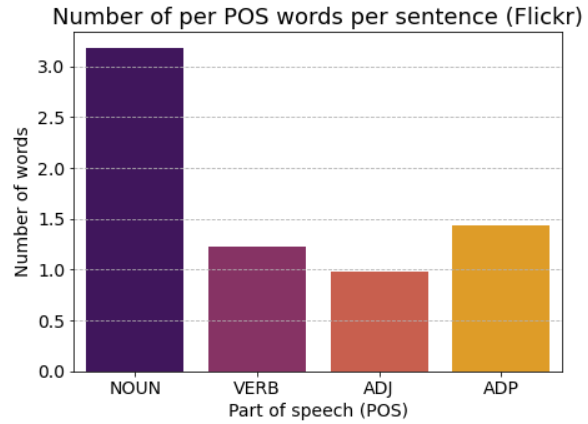}
    }
    % Second subplot
    \subfigure[Per dataset numbers.]{
        \includegraphics[width=0.4\textwidth]{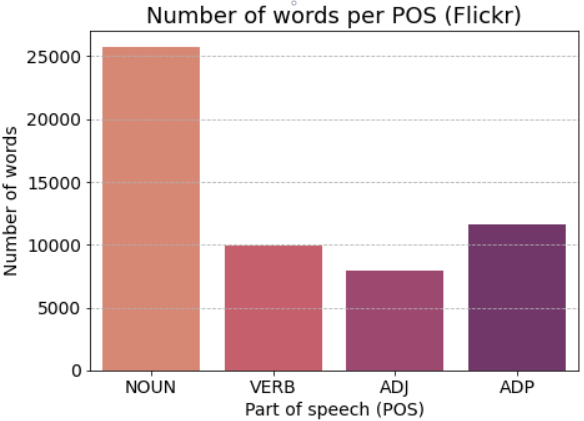}
    }
    \caption{Distributions of POS per each caption and overall in Flickr dataset.}
    \label{fig:two_plots}
\end{figure}
With an average of 3.181 nouns, 1.2249 verbs, 0.9817 adjectives and 1.4292 adpositions for Flickr sentences,  our queries are dense enough in terms of number of semantics to be perturbed; therefore, our contrastive edits are able to present an adequately divergent meaning compared to their ground truth counterparts.

POS distributions are illustrated in Figure \ref{fig:two_plots}.

\section{Contrastive edits examples}
\label{sec:contrastive-examples}
In the following Tables, \textbf{bold} words in 'Original' column denote concepts to be substituted, while \textbf{bold} words in the 'Contrastive' column denote correctly substituted concepts. Incorrect substitutions are underlined and colored in \textcolor{red}{red} in 'Contrastive' column (if none, then presented substitutions are correct). Correct \textbf{Blank (B)} substitutions have no bold, underlined or colored words. For \textbf{\textit{NOUNS}}, regarding \textbf{P} experiment, where singulars are exclusively substituted with singulars and plurals with plurals, valid substitutions are indicated with the \textbf{same} color in 'Original' column: \textbf{\textcolor{blue}{blue}} denotes singulars and \textbf{\textcolor{magenta}{magenta}} denotes plurals. The opposite happens in \textbf{SPS} experiments where singulars should be substituted with plurals and vice versa. Therefore, for \textbf{SPS} colored words in the 'Original' column should be substituted with words having different color: \textbf{\textcolor{PineGreen}{green}} denoting singulars should be substituted with \textbf{\textcolor{orange}{orange}}, which denotes plurals, and vice versa. (As for \textbf{RP} words are substituted irrespectively of their grammatical number, therefore candidates for substitutions are not colored.)
\begin{table*}[h!]
\caption{Examples for \textit{\textbf{ADP}} interventions on Flickr queries. }
\label{tab:examples-adp}
\centering \small 
\begin{tabular}{p{0.3cm}p{1cm}|p{16em}|p{16em}p{0em}}
\hline
 & Interv. & Original & Contrastive\\
\hline
\multirow{16}{30em}{\begin{turn}{90}
ADP \end{turn}} & \textbf{E} &  the boy \textbf{in} the blue shirt is swinging a baseball bat \textbf{towards} a ball \textbf{as} the boy \textbf{in} the red helmet waits to catch him out & the boy \textbf{on} the blue shirt is swinging a baseball bat \textbf{in} a ball \textbf{in} the boy \textbf{behind} the red helmet waits to catch him out \\ \cline{2-4}
& \textbf{SG-E} & the boy \textbf{in} the blue shirt is swinging a baseball bat towards a ball as the boy in the red helmet waits to catch him out & the boy \textbf{of} the blue shirt is swinging a baseball bat towards a ball as the boy in the red helmet waits to catch him out \\ \cline{2-4}
& \textbf{B} & the boy \textbf{in} the blue shirt is swinging a baseball bat \textbf{towards} a ball \textbf{as} the boy \textbf{in} the red helmet waits to catch him out & the boy the blue shirt is swinging a baseball bat a ball the boy the red helmet waits to catch him out \\\cline{2-4}
& \textbf{SG-B} & the boy \textbf{in} the blue shirt is swinging a baseball bat towards a ball as the boy in the red helmet waits to catch him out & the boy the blue shirt is swinging a baseball bat towards a ball as the boy in the red helmet waits to catch him out \\\cline{2-4}
\hline
\end{tabular}
\end{table*}
\begin{table*}[h!]
\centering \small
\caption{Examples for \textit{\textbf{ADJ}} and \textbf{\textit{NOUN }}interventions on Flickr queries.}
\label{tab:examples-adj}
\begin{tabular}{p{0.5cm}>{\centering\arraybackslash}p{4.3em}|p{15.7em}|p{15.7em}p{0em}}
\hline
POS &  Interv. & Original & Contrastive\\
\hline
\multirow{25}{0.1em}{\begin{turn}{90}
ADJ\end{turn}} & \textbf{CA} & a wet \textbf{black} dog is carrying a \textbf{green} toy through the grass & a wet \textbf{wheat} dog is carrying a \textbf{hotpink} toy through the grass \\\cline{2-4}
& \textbf{CI} & a wet \textbf{black} dog is carrying a \textbf{green} toy through the grass & a wet \textbf{tan} dog is carrying a \textbf{brown} toy through the grass \\\cline{2-4}
& \textbf{S} & a \textbf{little} girl in a pink dress going into a wooden cabin & a \textbf{big} girl in a pink dress going into a wooden cabin \\\cline{2-4}
& \textbf{A} & several \textbf{young} people sitting on a rail above a \textbf{crowded} beach & several \textbf{old} people sitting on a rail above a \textbf{uncrowded} beach \\\cline{2-4}
& \textbf{E} & two \textbf{young} people are approached by a \textbf{flamboyant} \textbf{young} woman dressed in a \textbf{red} bikini and a \textbf{red} feathered headdress & two \textbf{rundown} people are approached by a \textbf{silver} \textbf{rundown} woman dressed in a \textbf{skater} bikini and a \textbf{skater} feathered headdress \\\cline{2-4}
& \textbf{P} & two \textbf{young} people are approached by a \textbf{flamboyant} \textbf{young} woman dressed in a \textbf{red} bikini and a \textbf{red} feathered headdress & two \textbf{red} people are approached by a \textbf{red} \textbf{flamboyant} woman dressed in a \textbf{young} bikini and a \textbf{young} feathered headdress \\\cline{2-4}
& \textbf{B} & a \textbf{little} girl in a \textbf{pink} dress going into a \textbf{wooden} cabin & a girl in a dress going into a cabin \\\cline{2-4}
& \textbf{SG-CA} & a wet \textbf{black} dog is carrying a green toy through the grass & a wet \textbf{sandybrown} dog is carrying a green toy through the grass \\\cline{2-4}
& \textbf{SG-CI} & a wet \textbf{black} dog is carrying a green toy through the grass & a wet \textbf{chocolate} dog is carrying a green toy through the grass \\\cline{2-4}
& \textbf{SG-A} & several \textbf{young} people sitting on a rail above a crowded beach & several \textbf{old} people sitting on a rail above a crowded beach \\\cline{2-4}
& \textbf{SG-E} & two \textbf{young} people are approached by a flamboyant young woman dressed in a red bikini and a red feathered headdress & two \textbf{irish} people are approached by a flamboyant young woman dressed in a red bikini and a red feathered headdress \\\cline{2-4}
& \textbf{SG-B} & a \textbf{little} girl in a pink dress going into a wooden cabin & a girl in a pink dress going into a wooden cabin \\\cline{2-4}
\multirow{20}{30em}{\begin{turn}{90}
NOUN \end{turn}} 
& \textbf{E} & two \textbf{dogs} on \textbf{pavement} moving toward each other & two \textbf{tables} on \textbf{practice} moving toward each other \\\cline{2-4}
& \textbf{HE} & two \textbf{dogs} on \textbf{pavement} moving toward each other & two \textbf{canine} on \textbf{paved surface} moving toward each other \\\cline{2-4}
& \textbf{HO} & two \textbf{dogs} on \textbf{pavement} moving toward each other & two \textbf{basenji} on \textbf{curbside} moving toward each other \\\hline
& \textbf{P} & the \textbf{\textcolor{blue}{man}} with pierced \textbf{\textcolor{magenta}{ears}} is wearing \textbf{\textcolor{magenta}{glasses}} and an orange \textbf{\textcolor{blue}{hat}} & the \textbf{\textcolor{blue}{hat}} with pierced \textbf{\textcolor{magenta}{glasses}} is wearing \textbf{\textcolor{magenta}{ears}} and an orange \textbf{\textcolor{blue}{man}} \\\cline{2-4}
& \textbf{SPS} & the \textbf{\textcolor{PineGreen}{man}} with pierced \textbf{\textcolor{orange}{ears}} is wearing \textbf{\textcolor{orange}{glasses}} and an orange \textbf{\textcolor{PineGreen}{hat}} & the \textbf{\textcolor{orange}{ears}} with pierced \textbf{\textcolor{PineGreen}{man}} is wearing \textbf{\textcolor{PineGreen}{hat}} and an orange \textbf{\textcolor{orange}{glasses}} \\\cline{2-4}
& \textbf{RP} & the \textbf{man} with pierced \textbf{ears} is wearing \textbf{glasses} and an orange \textbf{hat} & the \textbf{hat} with pierced \textbf{glasses} is wearing \textbf{man} and an orange \textbf{ears} \\\cline{2-4}
& \textbf{B} & two \textbf{dogs} on \textbf{pavement} moving toward each other & two on moving toward each other \\\cline{2-4}
& \textbf{E-sing} &  two dogs on \textbf{pavement} moving toward each other & two \textbf{dogs} on \textbf{practice} moving toward each other \\ \cline{2-4}
& \textbf{B-sing} & two \textbf{dogs} on \textbf{pavement} moving toward each other & two dogs on moving toward each other \\\cline{2-4}
& \textbf{SG-E} & two \textbf{dogs} on pavement moving toward each other & two \textbf{teens} on pavement moving toward each other \\\cline{2-4}
& \textbf{SG-B} & two \textbf{dogs} on pavement moving toward each other & two on pavement moving toward each other \\
\hline 
\end{tabular}

\end{table*}

\end{document}

%% file: adj_table.tex
\begin{table*}[h]
\centering
\caption{Results ($\textit{ACE}_{R@1}$ metric) for \textbf{LR} on Flickr for ADJ POS singulars \& plurals. \textcolor{darkerdustygreen}{Green}/\textcolor{darkerdustypink}{Pink} denote \textcolor{darkerdustygreen}{more}/\textcolor{darkerdustypink}{less}  influential interventions.}
\label{tab:adj-table}
\begin{tabular}{p{0.3cm}c|>{\centering\arraybackslash}p{1.5cm}>{\centering\arraybackslash}p{1.5cm}>{\centering\arraybackslash}p{1cm}>{\centering\arraybackslash}p{1.5cm}>{\centering\arraybackslash}p{1.5cm}>{\centering\arraybackslash}p{1.1cm}>{\centering\arraybackslash}p{1.1cm}}
\toprule
&  & \multicolumn{7}{c}{\textbf{ADJ}} \\
\hline
& \small \textbf{Model} $M$ & \small \textbf{Color-all (CA)} & \small \textbf{Color-in (CI)} & \small \textbf{Size (S)} & \small \textbf{Antonym (A)} & \small \textbf{External (E)} & \small \textbf{Perm. (P)} & \small \textbf{Blank (B)}\\
\hline
\multirow{4}{0.4em}{\begin{turn}{90}
\small all - \end{turn}}&
\small distilroberta & \small 2.8 % color_all
& \small  3.06 % color_in
& \small 1.23 % sizes
& \small 2.96 % antonyms
& \small 2.05 % external_attribute
& \small \cellcolor{dustypink}{0.46} % attribute_permutation
& \small 1.64 % blank_attributes
\\
& \small MiniLM-L6 
& \small 2.87 % color_all
& \small  2.97 % color_in
& \small 1.43 % sizes
& \small 2.66 % antonyms
& \small 2.15 % external_attribute
& \small \cellcolor{dustypink}{0.23} % attribute_permutation
& \small 1.68 % blank_attributes
\\
& \small MiniLM-L12 & \small 3.03 % color_all
& \small  3.2 % color_in
& \small 1.39 % sizes
& \small 2.88 % antonyms
& \small 2.14 % external_attribute
& \small \cellcolor{dustypink}{0.29} % attribute_permutation
& \small 1.62 % blank_attributes
\\
& \small roberta-large & \small 2.93 % color_all
& \small  3.14 % color_in
& \small 1.66 % sizes
& \small 3.24 % antonyms
& \small 2.15 % external_attribute
& \small \cellcolor{dustypink}{0.44} % attribute_permutation
& \small 1.69 % blank_attributes
\\
\hline
\multirow{8}{0.4em}{\begin{turn}{90}
\small paraphrase - \end{turn}}&
\small MiniLM-L6 & \small 2.99 % color_all
& \small  3.3 % color_in
& \small 1.4 % sizes
& \small 3.11 % antonyms
& \small 2.37 % external_attribute
& \small \cellcolor{dustypink}{0.31} % attribute_permutation
& \small 1.87 % blank_attributes
\\
& \small MiniLM-L12 
& \small  3.01 % color_all
& \small 2.69 % color_in
& \small \cellcolor{dustypink}{0.42} % sizes
& \small 2.69 % antonyms
& \small 2.35 % external_attribute
& \small \cellcolor{dustypink}{0.1} % attribute_permutation
& \small 1.8 % blank_attributes
\\
& \small mpnet-base 
& \small 2.98 % color_all
& \small 3.2 % color_in
& \small \cellcolor{dustypink}{0.77} % sizes
& \small 2.97 % antonyms
& \small 2.06 % external_attribute
& \small \cellcolor{dustypink}{0.33} % attribute_permutation
& \small 1.76 % blank_attributes
\\
& \small albert-base
& \small 3.14 % color_all
& \small  3.26 % color_in
& \small 1.58 % sizes
& \small 3.18 % antonyms
& \small 2.16 % external_attribute
& \small \cellcolor{dustypink}{0.5} % attribute_permutation
& \small 1.88 % blank_attributes
\\
& \small albert-small 
& \small  3.33 % color_all
& \small 3.3 % color_in
& \small 1.7 % sizes
& \small 3.01 % antonyms
& \small 2.31 % external_attribute
& \small \cellcolor{dustypink}{0.2} % attribute_permutation
& \small 1.79 % blank_attributes
\\ 
& \small TinyBERT 
& \small 2.88 % color_all
& \small  3.14 % color_in
& \small \cellcolor{dustypink}{0.98} % sizes
& \small 2.72 % antonyms
& \small 2.29 % external_attribute
& \small \cellcolor{dustypink}{0.4} % attribute_permutation
& \small 1.7 % blank_attributes
\\ 
& \small distilroberta 
& \small 2.76 % color_all
& \small 2.86 % color_in
& \small 1.18 % sizes
& \small  3.09 % antonyms
& \small 2.33 % external_attribute
& \small \cellcolor{dustypink}{0.5} % attribute_permutation
& \small 1.71 % blank_attributes
\\ 
& \small XLM distil.
& \small 3.01 % color_all
& \small 3.06 % color_in
& \small 1.59 % sizes
& \small  \cellcolor{dustygreen}{6.5} % antonyms
& \small 2.26 % external_attribute
& \small \cellcolor{dustypink}{0.59} % attribute_permutation
& \small 1.83 % blank_attributes
\\ 
\hline
\multirow{4}{0.4em}{\begin{turn}{90}
\small stsb - \end{turn}}&
\small roberta-base 
& \small 2.93 % color_all
& \small 3.11 % color_in
& \small \cellcolor{dustypink}{0.93}% sizes
& \small  5.93 % antonyms
& \small 2.3 % external_attribute
& \small \cellcolor{dustypink}{0.44} % attribute_permutation
& \small 1.78 % blank_attributes
\\
& \small roberta-large 
& \small 2.24 % color_all
& \small 2.45 % color_in
& \small \cellcolor{dustypink}{0.75} % sizes
& \small  \cellcolor{dustygreen}{5.37} % antonyms
& \small 1.77 % external_attribute
& \small \cellcolor{dustypink}{0.18} % attribute_permutation
& \small 1.4 % blank_attributes
\\ 
& \small distilroberta 
& \small  3.12 % color_all
& \small 3.45 % color_in
& \small 1.21 % sizes
& \small 2.87 % antonyms
& \small 2.22 % external_attribute
& \small \cellcolor{dustypink}{0.2}% attribute_permutation
& \small 1.68 % blank_attributes
\\
& \small mpnet-base 
& \small 2.91 % color_all
& \small 3.14 % color_in
& \small 1.08 % sizes
& \small 2.62 % antonyms
& \small 2.37 % external_attribute
& \small \cellcolor{dustypink}{0.16} % attribute_permutation
& \small 1.67 % blank_attributes
\\ \hline
\multirow{3}{0.4em}{\begin{turn}{90}
\small multi-qa - \end{turn}}&
\small distilbert 
& \small 2.88 % color_all
& \small  3.23 % color_in
& \small 1.03 % sizes
& \small 3.72 % antonyms
& \small 2.17 % external_attribute
& \small \cellcolor{dustypink}{0.27} % attribute_permutation
& \small 1.62 % blank_attributes
\\
& \small mpnet-base 
& \small 4.59 % color_all
& \small \cellcolor{dustygreen}{4.62} % color_in
& \small 1.73 % sizes
& \small 1.32 % antonyms
& \small 2.92 % external_attribute
& \small \cellcolor{dustypink}{0.27} % attribute_permutation
& \small 2.07 % blank_attributes
\\ 
& \small MiniLM-L6 
& \small 2.76 % color_all
& \small 2.51 % color_in
& \small \cellcolor{dustypink}{0.74} % sizes
& \small 3.92 % antonyms
& \small 2.33 % external_attribute
& \small \cellcolor{dustypink}{0.2} % attribute_permutation
& \small 1.79 % blank_attributes
\\
\hline
\multirow{1}{0.4em}{\begin{turn}{90}
\small  \end{turn}}&
\small T5-gtr-base 
& \small 3.01 % color_all
& \small 2.69 % color_in
& \small \cellcolor{dustypink}{0.42} % sizes
& \small 2.39 % antonyms
& \small 2.35 % external_attribute
& \small \cellcolor{dustypink}{0.1} % attribute_permutation
& \small 1.8 % blank_attributes
\\ 
& \small nq-distilbert
& \small 2.84 % color_all
& \small 2.97 % color_in
& \small 1.76 % sizes
& \small \cellcolor{dustygreen}{7.64} % antonyms
& \small 1.98 % external_attribute
& \small \cellcolor{dustypink}{0.28} % attribute_permutation
& \small 1.53 % blank_attributes
\\
& \small nli-distilrob.
& \small 4.28 % color_all
& \small 4.37 % color_in
& \small 1.65 % sizes
& \small 2.27 % antonyms
& \small 2.66 % external_attribute
& \small \cellcolor{dustypink}{0.15} % attribute_permutation
& \small 2.1 % blank_attributes
\\ 
\hline
\end{tabular}
\end{table*}

%% file: adj_table_sing.tex
\begin{table*}[h]
\centering
\caption{Continuation of Table \ref{tab:adj-table}. Results ($\textit{ACE}_{R@1}$ metric) for \textbf{LR} on Flickr for ADJ POS, intervening on a single adjective at a time. \textcolor{darkerdustygreen}{Green}/\textcolor{darkerdustypink}{Pink} denote \textcolor{darkerdustygreen}{more}/\textcolor{darkerdustypink}{less} influential interventions.}
\label{tab:adj-table-sing}
\begin{tabular}{p{0.3cm}c|>{\centering\arraybackslash}p{1.8cm}>{\centering\arraybackslash}p{1.8cm}>{\centering\arraybackslash}p{1.8cm}>{\centering\arraybackslash}p{1.8cm}>{\centering\arraybackslash}p{1.8cm}>{\centering\arraybackslash}p{1.8cm}>{\centering\arraybackslash}p{0.8em}>{\centering\arraybackslash}p{0.8em}}
\toprule
&  & \multicolumn{5}{c}{\textbf{Single ADJ}} \\
\hline
& \small \textbf{Model} $M$ & \textbf{Color-all (SG-CA)} & \small \textbf{Color-in (SG-CI)} & \small \textbf{Antonym (SG-A)} & \small \textbf{External (SG-E)} & \small \textbf{Blank (SG-B)}\\
\hline
\multirow{4}{0.4em}{\begin{turn}{90}
\small all - \end{turn}}&
\small distilroberta 
& \small 3.52 % single_color_all
& \small 3.74 % single_color_in
& \small 2.01 % single_antonyms
& \small 1.92 % single_external_attribute
& \small 1.75 % single_blank_attributes
\\
& \small MiniLM-L6 
& \small 3.49 % single_color_all
& \small 3.47 % single_color_in
& \small 1.9 % single_antonyms
& \small 1.63 % single_external_attribute
& \small 1.71 % single_blank_attributes
\\
& \small MiniLM-L12 
& \small 3.65 % single_color_all
& \small 3.73 % single_color_in
& \small 2.0 % single_antonyms
& \small 1.78 % single_external_attribute
& \small 1.69 % single_blank_attributes
\\
& \small roberta-large 
& \small 3.6 % single_color_all
& \small 3.99 % single_color_in
& \small 2.3 % single_antonyms
& \small 2.1 % single_external_attribute
& \small 1.91 % single_blank_attributes
\\
\hline
\multirow{8}{0.4em}{\begin{turn}{90}
\small paraphrase - \end{turn}}&
\small MiniLM-L6 
& \small 3.54 % single_color_all
& \small 3.85 % single_color_in
& \small 2.25 % single_antonyms
& \small 2.23 % single_external_attribute
& \small 2.04 % single_blank_attributes
\\
& \small MiniLM-L12 
& \small 3.3 % single_color_all
& \small 3.14 % single_color_in
& \small 1.52 % single_antonyms
& \small 1.86 % single_external_attribute
& \small 1.58 % single_blank_attributes
\\
& \small mpnet-base 
& \small 3.56 % single_color_all
& \small 3.74 % single_color_in
& \small 2.07 % single_antonyms
& \small 2.0 % single_external_attribute
& \small 1.85 % single_blank_attributes
\\
& \small albert-base
& \small 3.64 % single_color_all
& \small 3.98 % single_color_in
& \small 2.3 % single_antonyms
& \small 2.24 % single_external_attribute
& \small 2.03 % single_blank_attributes
\\
& \small albert-small 
& \small \cellcolor{dustygreen}{4.06} % single_color_all
& \small 3.91 % single_color_in
& \small 2.17 % single_antonyms
& \small 1.92 % single_external_attribute
& \small 1.97 % single_blank_attributes
\\ 
& \small TinyBERT 
& \small 3.3 % single_color_all
& \small 3.65 % single_color_in
& \small 1.94 % single_antonyms
& \small 1.95 % single_external_attribute
& \small 1.72 % single_blank_attributes
\\ 
& \small distilroberta 
& \small 3.41 % single_color_all
& \small 3.56 % single_color_in
& \small 2.2 % single_antonyms
& \small 2.23 % single_external_attribute
& \small 1.77 % single_blank_attributes
\\ 
& \small XLM distil.
& \small 3.61 % single_color_all
& \small 3.57 % single_color_in
& \small 5.99 % single_antonyms
& \small 2.16 % single_external_attribute
& \small 2.01 % single_blank_attributes
\\ 
\hline
\multirow{4}{0.4em}{\begin{turn}{90}
\small stsb - \end{turn}}&
\small roberta-base 
& \small 3.47 % single_color_all
& \small 3.6 % single_color_in
& \small \cellcolor{dustygreen}{5.21} % single_antonyms
& \small 2.12 % single_external_attribute
& \small 1.9 % single_blank_attributes
\\
& \small roberta-large 
& \small 2.99 % single_color_all
& \small 2.92 % single_color_in
& \small \cellcolor{dustygreen}{4.46} % single_antonyms
& \small 1.68 % single_external_attribute
& \small 1.45 % single_blank_attributes
\\ 
& \small distilroberta 
& \small 3.69 % single_color_all
& \small \cellcolor{dustygreen}{4.02} % single_color_in
& \small 2.02 % single_antonyms
& \small 2.06 % single_external_attribute
& \small 1.72 % single_blank_attributes
\\
& \small mpnet-base 
& \small 3.54 % single_color_all
& \small 3.64 % single_color_in
& \small 1.84 % single_antonyms
& \small 1.95 % single_external_attribute
& \small 1.85 % single_blank_attributes
\\ \hline
\multirow{3}{0.4em}{\begin{turn}{90}
\small multi-qa - \end{turn}}&
\small distilbert 
& \small 3.51 % single_color_all
& \small 3.76 % single_color_in
& \small 2.82 % single_antonyms
& \small 1.95 % single_external_attribute
& \small 1.55 % single_blank_attributes
\\
& \small mpnet-base 
& \small \cellcolor{dustygreen}{5.97} % single_color_all
& \small \cellcolor{dustygreen}{5.84} % single_color_in
& \small 0.39 % single_antonyms
& \small 2.55 % single_external_attribute
& \small 2.2 % single_blank_attributes
\\ 
& \small MiniLM-L6 
& \small 3.11 % single_color_all
& \small 2.94 % single_color_in
& \small 2.84 % single_antonyms
& \small 1.85 % single_external_attribute
& \small 1.79 % single_blank_attributes
\\
\hline
\multirow{1}{0.4em}{\begin{turn}{90}
\small  \end{turn}}&
\small T5-gtr-base 
& \small 3.3 % single_color_all
& \small 3.14 % single_color_in
& \small 1.15 % single_antonyms
& \small 1.86 % single_external_attribute
& \small 1.58 % single_blank_attributes
\\ 
& \small nq-distilbert
& \small 3.3 % single_color_all
& \small 3.59 % single_color_in
& \small \cellcolor{dustygreen}{6.66} % single_antonyms
& \small 1.9 % single_external_attribute
& \small 1.69 % single_blank_attributes
\\
& \small nli-distilrob.
& \small \cellcolor{dustygreen}{4.9} % single_color_all
& \small \cellcolor{dustygreen}{4.96}% single_color_in
& \small 1.38 % single_antonyms
& \small 2.49 % single_external_attribute
& \small 2.13 % single_blank_attributes
\\ 
\hline
\end{tabular}
\end{table*}

%% file: adposition.tex
\begin{table*}[h]
\caption{Results ($\textit{ACE}_{R@1}$ metric) for \textbf{LR} on Flickr for ADP POS. \textcolor{darkerdustygreen}{Green}/\textcolor{darkerdustypink}{Pink} denote \textcolor{darkerdustygreen}{more}/\textcolor{darkerdustypink}{less} influential interventions.}
\label{tab:adposition}
\centering
\begin{tabular}{p{0.3cm}c|>{\centering\arraybackslash}p{2cm}>{\centering\arraybackslash}p{2cm}>{\centering\arraybackslash}p{2.6cm}>{\centering\arraybackslash}p{2.3cm}}
\hline
&  &  \multicolumn{4}{c}{\textbf{ADP}}  \\
\hline
& \small \textbf{Model} $M$ & \small \textbf{External} \newline \textbf{(E)} & \small \textbf{Blank} \newline \textbf{(B)} & \small \textbf{Single External (SG-E) } & \small \textbf{Single Blank (SG-B)} \\
\hline
\multirow{4}{0.4em}{\begin{turn}{90}
\small all - \end{turn}}&
\small distilroberta 
& \small \cellcolor{dustypink}{0.19} % external_adpositions
& \small \cellcolor{dustypink}{0.28} % blank_adpositions
& \small \cellcolor{dustypink}{0.33}% single_external_adpositions
& \small \cellcolor{dustypink}{0.26} % single_blank_adpositions
\\
& \small MiniLM-L6 
& \small \cellcolor{dustypink}{0.19} % external_adpositions
& \small \cellcolor{dustypink}{0.2} % blank_adpositions
& \small \cellcolor{dustypink}{0.25} % single_external_adpositions
& \small \cellcolor{dustypink}{0.19} % single_blank_adpositions
\\
& \small MiniLM-L12 
& \small \cellcolor{dustypink}{0.2} % external_adpositions
& \small \cellcolor{dustypink}{0.22} % blank_adpositions
& \small \cellcolor{dustypink}{0.34} % single_external_adpositions
& \small \cellcolor{dustypink}{0.29} % single_blank_adpositions
\\
& \small roberta-large
& \small \cellcolor{dustypink}{0.33} % external_adpositions
& \small \cellcolor{dustypink}{0.37} % blank_adpositions
& \small \cellcolor{dustypink}{0.45} % single_external_adpositions
& \small \cellcolor{dustypink}{0.33} % single_blank_adpositions
\\
\hline
\multirow{8}{0.4em}{\begin{turn}{90}
\small paraphrase - \end{turn}}&
\small MiniLM-L6 
& \small \cellcolor{dustypink}{0.26} % external_adpositions
& \small \cellcolor{dustypink}{0.21} % blank_adpositions
& \small \cellcolor{dustypink}{0.3} % single_external_adpositions
& \small \cellcolor{dustypink}{0.2} % single_blank_adpositions
\\
& \small MiniLM-L12 
& \small \cellcolor{dustypink}{0.07} % external_adpositions
& \small \cellcolor{dustypink}{0.09} % blank_adpositions
& \small \cellcolor{dustypink}{0.15} % single_external_adpositions
& \small \cellcolor{dustypink}{0.13} % single_blank_adpositions
\\
& \small mpnet-base 
& \small \cellcolor{dustypink}{0.28} % external_adpositions
& \small \cellcolor{dustypink}{0.18} % blank_adpositions
& \small \cellcolor{dustypink}{0.13} % single_external_adpositions
& \small \cellcolor{dustypink}{0.17} % single_blank_adpositions
\\
& \small albert-base
& \small \cellcolor{dustypink}{0.34} % external_adpositions
& \small \cellcolor{dustypink}{0.37} % blank_adpositions
& \small \cellcolor{dustypink}{0.41} % single_external_adpositions
& \small \cellcolor{dustypink}{0.39} % single_blank_adpositions
\\
& \small albert-small 
& \small \cellcolor{dustypink}{0.15} % external_adpositions
& \small \cellcolor{dustypink}{0.11} % blank_adpositions
& \small \cellcolor{dustypink}{0.12} % single_external_adpositions
& \small \cellcolor{dustypink}{0.09} % single_blank_adpositions
\\ 
& \small TinyBERT 
& \small \cellcolor{dustypink}{0.32} % external_adpositions
& \small \cellcolor{dustypink}{0.27} % blank_adpositions
& \small \cellcolor{dustypink}{0.4} % single_external_adpositions
& \small \cellcolor{dustypink}{0.25} % single_blank_adpositions
\\ 
& \small distilroberta 
& \small \cellcolor{dustypink}{0.49} % external_adpositions
& \small \cellcolor{dustypink}{0.49} % blank_adpositions
& \small \cellcolor{dustypink}{0.63} % single_external_adpositions
& \small \cellcolor{dustypink}{0.53} % single_blank_adpositions
\\ 
& \small XLM distil.
& \small \cellcolor{dustypink}{0.51} % external_adpositions
& \small \cellcolor{dustypink}{0.43} % blank_adpositions
& \small \cellcolor{dustypink}{0.59} % single_external_adpositions
& \small \cellcolor{dustypink}{0.41} % single_blank_adpositions
\\ 
\hline
\multirow{4}{0.4em}{\begin{turn}{90}
\small stsb - \end{turn}}&
\small roberta-base 
& \small \cellcolor{dustypink}{0.34} % external_adpositions
& \small \cellcolor{dustypink}{0.3} % blank_adpositions
& \small \cellcolor{dustypink}{0.38} % single_external_adpositions
& \small \cellcolor{dustypink}{0.33} % single_blank_adpositions
\\
& \small roberta-large 
& \small \cellcolor{dustypink}{0.23} % external_adpositions
& \small \cellcolor{dustypink}{0.22} % blank_adpositions
& \small \cellcolor{dustypink}{0.23} % single_external_adpositions
& \small \cellcolor{dustypink}{0.22} % single_blank_adpositions
\\ 
& \small distilroberta 
& \small \cellcolor{dustypink}{0.27} % external_adpositions
& \small \cellcolor{dustypink}{0.32} % blank_adpositions
& \small \cellcolor{dustypink}{0.28} % single_external_adpositions
& \small \cellcolor{dustypink}{0.25} % single_blank_adpositions
\\
& \small mpnet-base 
& \small \cellcolor{dustypink}{0.27} % external_adpositions
& \small \cellcolor{dustypink}{0.31} % blank_adpositions
& \small \cellcolor{dustypink}{0.3} % single_external_adpositions
& \small \cellcolor{dustypink}{0.27} % single_blank_adpositions
\\ \hline
\multirow{3}{0.4em}{\begin{turn}{90}
\small multi-qa - \end{turn}}&
\small distilbert 
& \small \cellcolor{dustypink}{0.27} % external_adpositions
& \small \cellcolor{dustypink}{0.23} % blank_adpositions
& \small \cellcolor{dustypink}{0.17} % single_external_adpositions
& \small \cellcolor{dustypink}{0.2} % single_blank_adpositions
\\
& \small mpnet-base 
& \small \cellcolor{dustypink}{0.35} % external_adpositions
& \small \cellcolor{dustypink}{0.28} % blank_adpositions
& \small \cellcolor{dustypink}{0.39} % single_external_adpositions
& \small \cellcolor{dustypink}{0.2} % single_blank_adpositions
\\ 
& \small MiniLM-L6 
& \small \cellcolor{dustypink}{0.27} % external_adpositions
& \small \cellcolor{dustypink}{0.32} % blank_adpositions
& \small \cellcolor{dustypink}{0.28} % single_external_adpositions
& \small \cellcolor{dustypink}{0.36} % single_blank_adpositions
\\
\hline
\multirow{1}{0.4em}{\begin{turn}{90}
\small  \end{turn}}&
\small T5-gtr-base 
& \small \cellcolor{dustypink}{0.07} % external_adpositions
& \small \cellcolor{dustypink}{0.09} % blank_adpositions
& \small \cellcolor{dustypink}{0.15} % single_external_adpositions
& \small \cellcolor{dustypink}{0.13} % single_blank_adpositions
\\ 
& \small nq-distilbert
& \small \cellcolor{dustypink}{0.23} % external_adpositions
& \small \cellcolor{dustypink}{0.23} % blank_adpositions
& \small \cellcolor{dustypink}{0.26} % single_external_adpositions
& \small \cellcolor{dustypink}{0.15} % single_blank_adpositions
\\
& \small nli-distilrob.
& \small \cellcolor{dustypink}{0.15} % external_adpositions
& \small \cellcolor{dustypink}{0.11} % blank_adpositions
& \small \cellcolor{dustypink}{0.12} % single_external_adpositions
& \small \cellcolor{dustypink}{0.07} % single_blank_adpositions
\\ 
\hline
\end{tabular}
\end{table*}

%% file: noun1.tex
\begin{table*}[h]
\centering
\caption{Results ($\textit{ACE}_{R@1}$ metric) for \textbf{LR} on Flickr focusing on NOUN POS. \textcolor{darkerdustygreen}{Green}/\textcolor{darkerdustypink}{Pink} denote \textcolor{darkerdustygreen}{more}/\textcolor{darkerdustypink}{less} influential interventions.}
\label{tab:noun1}
\begin{tabular}{p{0.3cm}c|>{\centering\arraybackslash}p{1.4cm}>{\centering\arraybackslash}p{1.7cm}>{\centering\arraybackslash}p{1.7cm}>{\centering\arraybackslash}p{1.3cm}>{\centering\arraybackslash}p{1.8cm}>{\centering\arraybackslash}p{1.3cm}}
\hline
&  & \multicolumn{6}{c}{\textbf{NOUN}}  \\
\hline
& \small \textbf{Model} $M$ & \small \textbf{External (E) } & \small \textbf{Hypernym (HE)} & \small \textbf{Hyponym (HO)} & \small \textbf{Permut. (P)} & \small \textbf{Sing.-plural (SPS)} & \small \textbf{Random Perm. (RP)}  \\
\hline
\multirow{4}{0.4em}{\begin{turn}{90}
\small all - \end{turn}}&
\small distilroberta 
& \small 2.85 % external_noun_all
& \small 2.5  % hypernyms_nouns % single_blank_adpositions \\
& \small 2.82 % hyponyms_nouns
& \small \cellcolor{dustypink}{0.4} % noun_permutations
& \small \cellcolor{dustypink}{0.51} % singular_plural_swap
& \small \cellcolor{dustypink}{0.33} % random_noun_permutations
\\
& \small MiniLM-L6 
& \small 2.86 % external_noun_all
& \small 2.63 % hypernyms_nouns
& \small 2.88 % hyponyms_nouns
& \small \cellcolor{dustypink}{0.18} % noun_permutations
& \small \cellcolor{dustypink}{0.2} % singular_plural_swap
& \small \cellcolor{dustypink}{0.15} % random_noun_permutations
\\
& \small MiniLM-L12
& \small 2.8 % external_noun_all
& \small 2.49 % hypernyms_nouns
& \small 2.71 % hyponyms_nouns
& \small \cellcolor{dustypink}{0.26} % noun_permutations
& \small \cellcolor{dustypink}{0.39} % singular_plural_swap
& \small \cellcolor{dustypink}{0.24} % random_noun_permutations
\\
& \small roberta-large
& \small 2.78 % external_noun_all
& \small 2.29 % hypernyms_nouns
&  \small 2.51 % hyponyms_nouns
& \small \cellcolor{dustypink}{0.9} % noun_permutations
& \small 1.05 % singular_plural_swap
& \small \cellcolor{dustypink}{0.74} % random_noun_permutations
\\
\hline
\multirow{8}{0.4em}{\begin{turn}{90}
\small paraphrase - \end{turn}}&
\small MiniLM-L6 
& \small 3.03 % external_noun_all
& \small 2.49 % hypernyms_nouns
& \small 3.02 % hyponyms_nouns
& \small \cellcolor{dustypink}{0.44} % noun_permutations
& \small \cellcolor{dustypink}{0.56} % singular_plural_swap
& \small \cellcolor{dustypink}{0.37} % random_noun_permutations
\\
& \small MiniLM-L12 
& \small 3.01 % external_noun_all
& \small 2.67 % hypernyms_nouns
& \small 2.98 % hyponyms_nouns
& \small \cellcolor{dustypink}{0.18} % noun_permutations
& \small \cellcolor{dustypink}{0.24} % singular_plural_swap
& \small \cellcolor{dustypink}{0.1} % random_noun_permutations
\\
& \small mpnet-base 
& \small 2.92 % external_noun_all
& \small 2.35 % hypernyms_nouns
& \small 2.65 % hyponyms_nouns
& \small \cellcolor{dustypink}{0.48} % noun_permutations
& \small \cellcolor{dustypink}{0.72} % singular_plural_swap
& \small \cellcolor{dustypink}{0.44} % random_noun_permutations
\\
& \small albert-base
& \small 3.03 % external_noun_all
& \small 2.44 % hypernyms_nouns
& \small 2.92 % hyponyms_nouns
& \small \cellcolor{dustypink}{0.69} % noun_permutations
& \small \cellcolor{dustypink}{0.97} % singular_plural_swap
& \small \cellcolor{dustypink}{0.63} % random_noun_permutations
\\
& \small albert-small 
& \small 2.9 % external_noun_all
& \small 2.6 % hypernyms_nouns
& \small 3.03 % hyponyms_nouns
& \small \cellcolor{dustypink}{0.09} % noun_permutations
& \small \cellcolor{dustypink}{0.08} % singular_plural_swap
& \small \cellcolor{dustypink}{0.07} % random_noun_permutations
\\ 
& \small TinyBERT 
& \small 3.05 % external_noun_all
& \small 2.68 % hypernyms_nouns
& \small 2.86 % hyponyms_nouns
& \small \cellcolor{dustypink}{0.62} % noun_permutations
& \small \cellcolor{dustypink}{0.91} % singular_plural_swap
& \small \cellcolor{dustypink}{0.59} % random_noun_permutations
\\ 
& \small distilroberta 
& \small 3.18 % external_noun_all
& \small 2.74 % hypernyms_nouns
& \small 2.98 % hyponyms_nouns
& \small \cellcolor{dustypink}{0.88} % noun_permutations
& \small 1.26 % singular_plural_swap
& \small \cellcolor{dustypink}{0.76} % random_noun_permutations
\\ 
& \small XLM distil.
& \small 3.09 % external_noun_all
& \small 2.55 % hypernyms_nouns
& \small 2.77 % hyponyms_nouns
& \small \cellcolor{dustypink}{0.82} % noun_permutations
& \small 1.34 % singular_plural_swap
& \small \cellcolor{dustypink}{0.74} % random_noun_permutations
\\ 
\hline
\multirow{4}{0.4em}{\begin{turn}{90}
\small stsb - \end{turn}}&
\small roberta-base 
& \small 2.93 % external_noun_all
& \small 2.42 % hypernyms_nouns
& \small 2.66 % hyponyms_nouns
& \small \cellcolor{dustypink}{0.71} % noun_permutations
& \small \cellcolor{dustypink}{0.97} % singular_plural_swap
& \small \cellcolor{dustypink}{0.62} % random_noun_permutations
\\
& \small roberta-large 
& \small 3.09 % external_noun_all
& \small 2.42 % hypernyms_nouns
& \small 2.73 % hyponyms_nouns
& \small 1.0 % noun_permutations
& \small 1.72 % singular_plural_swap
& \small \cellcolor{dustypink}{0.82} % random_noun_permutations
\\ 
& \small distilroberta 
& \small 2.82 % external_noun_all
& \small 2.45 % hypernyms_nouns
& \small 2.65 % hyponyms_nouns
& \small \cellcolor{dustypink}{0.45} % noun_permutations
& \small \cellcolor{dustypink}{0.8} % singular_plural_swap
& \small \cellcolor{dustypink}{0.38} % random_noun_permutations
\\
& \small mpnet-base 
& \small 2.82 % external_noun_all
& \small 2.33 % hypernyms_nouns
& \small 2.6 % hyponyms_nouns
& \small \cellcolor{dustypink}{0.34} % noun_permutations
& \small \cellcolor{dustypink}{0.42} % singular_plural_swap
& \small \cellcolor{dustypink}{0.25} % random_noun_permutations
\\ \hline
\multirow{3}{0.4em}{\begin{turn}{90}
\small multi-qa - \end{turn}}&
\small distilbert 
& \small 3.11 % external_noun_all
& \small 2.72 % hypernyms_nouns
& \small 2.87 % hyponyms_nouns
& \small \cellcolor{dustypink}{0.62} % noun_permutations
& \small 1.25 % singular_plural_swap
& \small \cellcolor{dustypink}{0.5} % random_noun_permutations
\\
& \small mpnet-base 
& \small 2.85 % external_noun_all
& \small 2.32 % hypernyms_nouns
& \small 2.52 % hyponyms_nouns
& \small \cellcolor{dustypink}{0.09} % noun_permutations
& \small \cellcolor{dustypink}{0.51} % singular_plural_swap
& \small \cellcolor{dustypink}{0.05} % random_noun_permutations
\\ 
& \small MiniLM-L6 
& \small 3.14 % external_noun_all
& \small 2.81 % hypernyms_nouns
& \small 3.14 % hyponyms_nouns
& \small \cellcolor{dustypink}{0.61} % noun_permutations
& \small 1.26 % singular_plural_swap
& \small \cellcolor{dustypink}{0.46} % random_noun_permutations
\\
\hline
\multirow{1}{0.4em}{\begin{turn}{90}
\small  \end{turn}}&
\small T5-gtr-base 
& \small 2.68 % external_noun_all
& \small 2.4 % hypernyms_nouns
& \small 2.29 % hyponyms_nouns
& \small \cellcolor{dustypink}{0.2} % noun_permutations
& \small \cellcolor{dustypink}{0.6} % singular_plural_swap
& \small \cellcolor{dustypink}{0.14} % random_noun_permutations
\\ 
& \small nq-distilbert
& \small 3.52 % external_noun_all
& \small 3.14 % hypernyms_nouns
& \small 3.62 % hyponyms_nouns
& \small 1.96 % noun_permutations
& \small 4.12 % singular_plural_swap
& \small 1.55 % random_noun_permutations
\\
& \small nli-distilrob.
& \small 2.77 % external_noun_all
& \small 2.74 % hypernyms_nouns
 & \small 2.83 % hyponyms_nouns
& \small \cellcolor{dustypink}{0.22} % noun_permutations
& \small \cellcolor{dustypink}{0.26} % singular_plural_swap
& \small \cellcolor{dustypink}{0.2} % random_noun_permutations
\\ 
\hline
\end{tabular}
\end{table*}

%% file: noun2.tex
\begin{table}[h!]
    \centering
\caption{Continuation of Table \ref{tab:noun1}. Results ($\textit{ACE}_{R@1}$ metric) for \textbf{LR} on Flickr focusing on NOUN POS. \textcolor{darkerdustygreen}{Green}/\textcolor{darkerdustypink}{Pink} denote \textcolor{darkerdustygreen}{more}/\textcolor{darkerdustypink}{less} influential interventions.}
\label{tab:noun2}
    \begin{tabular}{p{0.3cm}c|>{\centering\arraybackslash}p{1.3cm}>{\centering\arraybackslash}p{2.2cm}>{\centering\arraybackslash}p{2.2cm}>{\centering\arraybackslash}p{1.8cm}>{\centering\arraybackslash}p{1.8cm}}
\hline
&  & \multicolumn{5}{c}{\textbf{NOUN}}  \\
\hline
& \small \textbf{Model} $M$ & \small \textbf{Blank (B)} & \small \textbf{External sing. (E-sing)} & \small \textbf{Blank sing.  (B-sing)} &\small \textbf{Single External (SG-E)} & \small \textbf{Single Blank (SG-B)} \\
\hline
\multirow{4}{0.4em}{\begin{turn}{90}
\small all - \end{turn}}&
\small distilroberta 
& \small 3.2 % blank_nouns
& \small \cellcolor{dustygreen}{4.19} % external_noun_singular
& \small 3.2 % blank_nouns_singular
& \small 3.19 % single_external_noun_all
& \small 3.02 % single_blank_nouns
\\
& \small MiniLM-L6 
& \small 3.19 % blank_nouns
& \small \cellcolor{dustygreen}{4.01} % external_noun_singular
& \small 3.21 % blank_nouns_singular
& \small 3.14 % single_external_noun_all
& \small 3.1 % single_blank_nouns
\\
& \small MiniLM-L12
& \small 3.21 % blank_nouns
& \small \cellcolor{dustygreen}{4.03} % external_noun_singular
& \small 3.2 % blank_nouns_singular
& \small 3.02 % single_external_noun_all
& \small 2.92 % single_blank_nouns
\\
& \small roberta-large
& \small 3.2 % blank_nouns
& \small \cellcolor{dustygreen}{4.04} % external_noun_singular
& \small 3.17 % blank_nouns_singular
& \small 3.33 % single_external_noun_all
& \small 2.88 % single_blank_nouns
\\
\hline
\multirow{8}{0.4em}{\begin{turn}{90}
\small paraphrase - \end{turn}}&
\small MiniLM-L6 
& \small 3.28 % blank_nouns
& \small \cellcolor{dustygreen}{4.32} % external_noun_singular
& \small 3.31 % blank_nouns_singular
& \small 3.74 % single_external_noun_all
& \small 3.54 % single_blank_nouns
\\
& \small MiniLM-L12 
& \small 3.21 % blank_nouns
& \small \cellcolor{dustygreen}{4.21} % external_noun_singular
& \small 3.27 % blank_nouns_singular
& \small 3.25 % single_external_noun_all
& \small 3.19 % single_blank_nouns
\\
& \small mpnet-base 
& \small 3.16 % blank_nouns
& \small \cellcolor{dustygreen}{4.15} % external_noun_singular
& \small 3.1 % blank_nouns_singular
& \small 3.44 % single_external_noun_all
& \small 3.11 % single_blank_nouns
\\
& \small albert-base
& \small 3.25 % blank_nouns
& \small \cellcolor{dustygreen}{4.28} % external_noun_singular
& \small 3.23 % blank_nouns_singular
& \small 3.54 % single_external_noun_all
& \small 3.1 % single_blank_nouns
\\
& \small albert-small 
& \small 3.27 % blank_nouns
& \small \cellcolor{dustygreen}{4.21} % external_noun_singular
& \small 3.33 % blank_nouns_singular
& \small 3.64 % single_external_noun_all
& \small 3.63 % single_blank_nouns
\\ 
& \small TinyBERT 
& \small 3.34 % blank_nouns
& \small \cellcolor{dustygreen}{4.36} % external_noun_singular
& \small 3.38 % blank_nouns_singular
& \small \cellcolor{dustygreen}{3.69} % single_external_noun_all
& \small 3.61 % single_blank_nouns
\\ 
& \small distilroberta 
& \small 3.38 % blank_nouns
& \small \cellcolor{dustygreen}{4.36} % external_noun_singular
& \small 3.36 % blank_nouns_singular
& \small 4.02 % single_external_noun_all
& \small 3.58 % single_blank_nouns
\\ 
& \small XLM distil.
& \small 3.25 % blank_nouns
& \small \cellcolor{dustygreen}{4.35} % external_noun_singular
& \small 3.22 % blank_nouns_singular
& \small 3.83 % single_external_noun_all
& \small 3.2 % single_blank_nouns
\\ 
\hline
\multirow{4}{0.4em}{\begin{turn}{90}
\small stsb - \end{turn}}&
\small roberta-base 
& \small 3.18 % blank_nouns
& \small \cellcolor{dustygreen}{4.25} % external_noun_singular
& \small 3.16 % blank_nouns_singular
& \small 3.39 % single_external_noun_all
& \small 3.12 % single_blank_nouns
\\
& \small roberta-large 
& \small 3.42 % blank_nouns
& \small \cellcolor{dustygreen}{4.3} % external_noun_singular
& \small 3.56 % blank_nouns_singular
& \small \cellcolor{dustygreen}{4.73} % single_external_noun_all
& \small \cellcolor{dustygreen}{4.82} % single_blank_nouns
\\ 
& \small distilroberta 
& \small 3.25 % blank_nouns
& \small \cellcolor{dustygreen}{4.09} % external_noun_singular
& \small 3.27 % blank_nouns_singular
& \small 3.65 % single_external_noun_all
& \small 3.74 % single_blank_nouns
\\
& \small mpnet-base 
& \small 3.12 % blank_nouns
& \small 3.94 % external_noun_singular
& \small 3.09 % blank_nouns_singular
& \small 3.34 % single_external_noun_all
& \small 3.07 % single_blank_nouns
\\ \hline
\multirow{3}{0.4em}{\begin{turn}{90}
\small multi-qa - \end{turn}}&
\small distilbert 
& \small 3.27 % blank_nouns
& \small \cellcolor{dustygreen}{4.28} % external_noun_singular
& \small 3.35 % blank_nouns_singular
& \small \cellcolor{dustygreen}{4.06} % single_external_noun_all
& \small 3.59 % single_blank_nouns
\\
& \small mpnet-base 
& \small 3.06 % blank_nouns
& \small \cellcolor{dustygreen}{4.01} % external_noun_singular
& \small 2.93 % blank_nouns_singular
& \small 2.47 % single_external_noun_all
& \small 2.03 % single_blank_nouns
\\ 
& \small MiniLM-L6 
& \small 3.29 % blank_nouns
& \small \cellcolor{dustygreen}{4.35} % external_noun_singular
& \small 3.45 % blank_nouns_singular
& \small \cellcolor{dustygreen}{4.1} % single_external_noun_all
& \small 3.95 % single_blank_nouns
\\
\hline
\multirow{1}{0.4em}{\begin{turn}{90}
\small  \end{turn}}&
\small T5-gtr-base 
& \small 3.01 % blank_nouns
& \small 3.81 % external_noun_singular
& \small 2.91 % blank_nouns_singular
& \small 1.91 % single_external_noun_all
& \small 1.86 % single_blank_nouns
\\ 
& \small nq-distilbert
& \small 3.51 % blank_nouns
& \small \cellcolor{dustygreen}{4.78} % external_noun_singular
& \small 3.97 % blank_nouns_singular
& \small \cellcolor{dustygreen}{6.49} % single_external_noun_all
& \small \cellcolor{dustygreen}{5.86} % single_blank_nouns
\\
& \small nli-distilrob.
& \small 3.18 % blank_nouns
& \small \cellcolor{dustygreen}{4.08} % external_noun_singular
& \small 3.17 % blank_nouns_singular
& \small 3.47 % single_external_noun_all
& \small 3.14 % single_blank_nouns
\\ 
\hline
    \end{tabular}
\end{table}

%% file: verb1.tex
\begin{table}[h!]
    \centering
\caption{(Results ($ACE_{R@1}$ metric) for \textbf{LR} on Flickr  focusing on VERB POS. \textcolor{darkerdustygreen}{Green}/\textcolor{darkerdustypink}{Pink} denote \textcolor{darkerdustygreen}{more}/\textcolor{darkerdustypink}{less} influential interventions.}
\label{tab:verb1}
    \begin{tabular}{p{0.3cm}c|>{\centering\arraybackslash}p{1.2cm}>{\centering\arraybackslash}p{2cm}>{\centering\arraybackslash}p{1.8cm}>{\centering\arraybackslash}p{1.6cm}>{\centering\arraybackslash}p{1.6cm}>{\centering\arraybackslash}p{1.6cm}}
\hline
&   &  \multicolumn{6}{c}{\textbf{VERB}} \\
\hline
& \small Model $M$  & \small \textbf{External (E)} & \small \textbf{External combined (E-comb)} & \small \textbf{Hypernym (HE)} & \small \textbf{Hyponym (HO)} & \small \textbf{Antonym (A)} & \small \textbf{Permut. (P)}  \\
\hline
\multirow{4}{0.4em}{\begin{turn}{90}
\small all - \end{turn}}&
\small distilroberta 
& \small 2.91 % external_verb_all
& \small \cellcolor{dustygreen}{4.0} % external_verb_combine
& \small 1.3 % hypernyms_verbs
& \small \cellcolor{dustygreen}{4.83} % hyponyms_verbs
& \small 2.22 % verb_antonyms
& \small \cellcolor{dustypink}{0.16}% verb_permutation
\\
& \small MiniLM-L6 
& \small 2.83 % external_verb_all
& \small \cellcolor{dustygreen}{3.95} % external_verb_combine
& \small 1.33 % hypernyms_verbs
& \small \cellcolor{dustygreen}{5.46} % hyponyms_verbs
& \small 2.13 % verb_antonyms
& \small \cellcolor{dustypink}{0.11} % verb_permutations
\\
& \small MiniLM-L12 
& \small 2.83 % external_verb_all
& \small \cellcolor{dustygreen}{4.08} % external_verb_combine
& \small 1.29 % hypernyms_verbs
& \small \cellcolor{dustygreen}{5.35} % hyponyms_verbs
& \small 2.15 % verb_antonyms
& \small \cellcolor{dustypink}{0.23} % verb_permutations
\\
& \small roberta-large
& \small 3.44 % external_verb_all
& \small \cellcolor{dustygreen}{4.05} % external_verb_combine
& \small 1.39 % hypernyms_verbs
& \small \cellcolor{dustygreen}{4.28} % hyponyms_verbs
& \small 2.28 % verb_antonyms
& \small \cellcolor{dustypink}{0.47} % verb_permutations
\\\hline
\multirow{8}{0.4em}{\begin{turn}{90}
\small paraphrase - \end{turn}}&
\small MiniLM-L6 
& \small 3.33 % external_verb_all
& \small \cellcolor{dustygreen}{4.31} % external_verb_combine
& \small 1.52 % hypernyms_verbs
& \small \cellcolor{dustygreen}{6.35} % hyponyms_verbs
& \small 2.54 % verb_antonyms
& \small \cellcolor{dustypink}{0.25} % verb_permutations
\\
& \small MiniLM-L12 
& \small 2.38 % external_verb_all
& \small 3.21 % external_verb_combine
& \small 1.09 % hypernyms_verbs
& \small \cellcolor{dustygreen}{4.85} % hyponyms_verbs
& \small 1.34 % verb_antonyms
& \small \cellcolor{dustypink}{0.01} % verb_permutations
\\
& \small mpnet-base 
& \small 3.18 % external_verb_all
& \small \cellcolor{dustygreen}{4.25} % external_verb_combine
& \small 1.32 % hypernyms_verbs
& \small \cellcolor{dustygreen}{3.97} % hyponyms_verbs
& \small 2.37 % verb_antonyms
& \small \cellcolor{dustypink}{0.25} % verb_permutations
\\
& \small albert-base 
& \small 3.12 % external_verb_all
& \small \cellcolor{dustygreen}{4.19} % external_verb_combine
& \small 1.33 % hypernyms_verbs
& \small \cellcolor{dustygreen}{5.49} % hyponyms_verbs
& \small 2.71 % verb_antonyms
& \small \cellcolor{dustypink}{0.42} % verb_permutations
\\
& \small albert-small 
& \small 3.35 % external_verb_all
& \small \cellcolor{dustygreen}{4.18} % external_verb_combine
& \small 1.57 % hypernyms_verbs
& \small \cellcolor{dustygreen}{5.82} % hyponyms_verbs
& \small 2.51 % verb_antonyms
& \small \cellcolor{dustypink}{0.1} % verb_permutations
\\
& \small TinyBERT 
& \small 3.34 % external_verb_all
& \small\cellcolor{dustygreen}{4.21} % external_verb_combine
& \small \cellcolor{dustypink}{0.38} % hypernyms_verbs
& \small \cellcolor{dustypink}{0.76} % hyponyms_verbs
& \small 2.49 % verb_antonyms
& \small \cellcolor{dustypink}{0.3} % verb_permutations
\\
& \small distilroberta 
& \small \cellcolor{dustygreen}{4.08} % external_verb_all
& \small \cellcolor{dustygreen}{4.65}% external_verb_combine
& \small 1.49 % hypernyms_verbs
& \small \cellcolor{dustygreen}{5.61} % hyponyms_verbs
& \small 3.01 % verb_antonyms
& \small \cellcolor{dustypink}{0.62} % verb_permutations
\\ 
& \small XLM distil. 
& \small 3.89 % external_verb_all
& \small \cellcolor{dustygreen}{4.59} % external_verb_combine
& \small 3.93 % hypernyms_verbs
& \small \cellcolor{dustygreen}{15.3} % hyponyms_verbs
& \small 3.02 % verb_antonyms
& \small \cellcolor{dustypink}{0.66} % verb_permutations
\\
\hline
\multirow{4}{0.4em}{\begin{turn}{90}
\small stsb - \end{turn}}&
\small roberta-base 
& \small \cellcolor{dustygreen}{9.81} % external_verb_all
& \small \cellcolor{dustygreen}{6.47} % external_verb_combine
& \small 3.26 % hypernyms_verbs
& \small \cellcolor{dustygreen}{12.8} % hyponyms_verbs
& \small 2.47 % verb_antonyms
& \small \cellcolor{dustygreen}{4.52} % verb_permutations
\\
& \small roberta-large 
& \small \cellcolor{dustygreen}{8.16} % external_verb_all
& \small \cellcolor{dustygreen}{6.83} % external_verb_combine
& \small 3.08 % hypernyms_verbs
& \small \cellcolor{dustygreen}{9.66} % hyponyms_verbs
& \small 2.38 % verb_antonyms
& \small 2.52 % verb_permutations
\\ 
& \small distilroberta 
& \small \cellcolor{dustygreen}{4.05} % external_verb_all
& \small \cellcolor{dustygreen}{4.4} % external_verb_combine
& \small 1.5 % hypernyms_verbs
& \small \cellcolor{dustygreen}{5.59} % hyponyms_verbs
& \small 2.2 % verb_antonyms
& \small \cellcolor{dustypink}{0.8} % verb_permutations
\\ 
& \small mpnet-base
& \small 3.9 % external_verb_all
& \small \cellcolor{dustygreen}{4.69} % external_verb_combine
& \small 1.51 % hypernyms_verbs
& \small \cellcolor{dustygreen}{4.36} % hyponyms_verbs
& \small 2.42 % verb_antonyms
& \small \cellcolor{dustypink}{0.27} % verb_permutations
\\ \hline
\multirow{3}{0.4em}{\begin{turn}{90}
\small multi-qa - \end{turn}}&
\small distilbert 
& \small \cellcolor{dustygreen}{4.92} % external_verb_all
& \small \cellcolor{dustygreen}{4.38} % external_verb_combine
& \small 1.84 % hypernyms_verbs
& \small \cellcolor{dustygreen}{7.42} % hyponyms_verbs
& \small 2.14 % verb_antonyms
& \small 1.8 % verb_permutations
\\
& \small mpnet-base 
& \small \cellcolor{dustypink}{0.28} % external_verb_all
& \small 2.4 % external_verb_combine
& \small \cellcolor{dustypink}{0.17} % hypernyms_verbs
& \small \cellcolor{dustypink}{0.87} % hyponyms_verbs
& \small 3.82 % verb_antonyms
& \small 1.48 % verb_permutations
\\ 
& \small MiniLM-L6 
& \small \cellcolor{dustygreen}{4.83} % external_verb_all
& \small \cellcolor{dustygreen}{4.17} % external_verb_combine
& \small 1.9 % hypernyms_verbs
& \small \cellcolor{dustygreen}{7.93} % hyponyms_verbs
& \small 1.49 % verb_antonyms
& \small 1.68 % verb_permutations
\\
\hline
\multirow{1}{0.4em}{\begin{turn}{90}
\small  \end{turn}}
& \small T5-gtr-base 
& \small \cellcolor{dustypink}{0.77} % external_verb_all
& \small 2.35 % external_verb_combine
& \small \cellcolor{dustypink}{0.42} % hypernyms_verbs
& \small 2.2 % hyponyms_verbs
& \small 1.34 % verb_antonyms
& \small 1.39 % verb_permutations
\\ 
& \small nq-distilbert
& \small \cellcolor{dustygreen}{12.43} % external_verb_all
& \small \cellcolor{dustygreen}{7.92} % external_verb_combine
& \small \cellcolor{dustygreen}{4.33} % hypernyms_verbs
& \small \cellcolor{dustygreen}{17.4} % hyponyms_verbs
& \small 2.37 % verb_antonyms
& \small \cellcolor{dustygreen}{6.46} % verb_permutations
\\
& \small nli-distilrob.
& \small 2.95 % external_verb_all
& \small \cellcolor{dustygreen}{4.5} % external_verb_combine
& \small 1.21 % hypernyms_verbs
& \small \cellcolor{dustygreen}{4.77} % hyponyms_verbs
& \small 1.38 % verb_antonyms
& \small \cellcolor{dustypink}{0.36} % verb_permutations
\\ 
\hline
    \end{tabular}
\end{table}

%% file: verb2.tex
\begin{table*}[t!]
\centering
\caption{Continuation of Table \ref{tab:verb1}. Results ($ACE_{R@1}$ metric for \textbf{LR} on Flickr  focusing on VERB POS. }
\label{tab:verb2}
\begin{tabular}{p{0.3cm}c|>{\centering\arraybackslash}p{1.2cm}>{\centering\arraybackslash}p{2.7cm}>{\centering\arraybackslash}p{2.7cm}>{\centering\arraybackslash}p{2.7cm}}
\hline
&   &  \multicolumn{4}{c}{\textbf{VERB}} \\
\hline
& \small Model $M$  & \small \textbf{Blank (B)} & \small \textbf{Single antonym (SG-A)} & \small \textbf{Single external (SG-E)} & \small \textbf{Single blank (SG-B)} \\
\hline
\multirow{4}{0.1em}{\begin{turn}{90}
\small all - \end{turn}}&
\small distilroberta 
& \small 1.91 % blank_verbs
& \small 2.21 % single_verb_antonyms
& \small \cellcolor{dustypink}{0.77} % single_external_verb_all
& \small 1.85 % single_blank_verbs
\\ 
% single_blank_verbs\\ 
& \small MiniLM-L6 
& \small 1.98 % blank_verbs
& \small 2.3 % single_verb_antonyms
& \small \cellcolor{dustypink}{0.8} % single_external_verb_all
& \small 1.95 % single_blank_verbs
\\
& \small MiniLM-L12 
& \small 1.98 % blank_verbs
& \small 2.24 % single_verb_antonyms
& \small \cellcolor{dustypink}{0.71} % single_external_verb_all
& \small 1.88 % single_blank_verbs
\\
& \small roberta-large
& \small 1.88 % blank_verbs
& \small 2.35 % single_verb_antonyms
& \small \cellcolor{dustypink}{0.99} % single_external_verb_all
& \small 1.88 % single_blank_verbs
\\
\hline
\multirow{8}{0.5em}{\begin{turn}{90}
\small paraphrase - \end{turn}}&
\small MiniLM-L6 
& \small 1.88 % blank_verbs
& \small 2.66 % single_verb_antonyms
& \small 1.01 % single_external_verb_all
& \small 1.92 % single_blank_verbs
\\
& \small MiniLM-L12 
& \small 1.56 % blank_verbs
& \small 1.43 % single_verb_antonyms
& \small \cellcolor{dustypink}{0.76} % single_external_verb_all
& \small 1.47 % single_blank_verbs
\\
& \small mpnet-base 
& \small 1.91 % blank_verbs
& \small 2.52 % single_verb_antonyms
& \small \cellcolor{dustypink}{0.99} % single_external_verb_all
& \small 1.93 % single_blank_verbs
\\
& \small albert-base 
& \small 1.91 % blank_verbs
& \small 2.87 % single_verb_antonyms
& \small 1.0 % single_external_verb_all
& \small 1.87 % single_blank_verbs
\\
& \small albert-small 
& \small 1.9 % blank_verbs
& \small 2.57 % single_verb_antonyms
& \small 0.93 % single_external_verb_all
& \small 1.9 % single_blank_verbs
\\ 

& \small TinyBERT 
& \small 1.89 % blank_verbs
& \small 2.56 % single_verb_antonyms
& \small 1.0 % single_external_verb_all
& \small 1.74 % single_blank_verbs
\\
& \small distilroberta 
& \small 2.15 % blank_verbs
& \small 3.17 % single_verb_antonyms
& \small 1.11 % single_external_verb_all
& \small 2.16 % single_blank_verbs
\\ 
& \small XLM distil. 
& \small 2.03 % blank_verbs
& \small 3.13 % single_verb_antonyms
& \small 1.0 % single_external_verb_all
& \small 1.98 % single_blank_verbs
\\
\hline
\multirow{4}{0.5em}{\begin{turn}{90}
\small stsb - \end{turn}}&
\small roberta-base 
& \small 1.84 % blank_verbs
& \small 2.51 % single_verb_antonyms
& \small 1.03 % single_external_verb_all
& \small 1.79 % single_blank_verbs
\\
& \small roberta-large 
& \small 1.9 % blank_verbs
& \small 2.57 % single_verb_antonyms
& \small \cellcolor{dustypink}{0.91} % single_external_verb_all
& \small 1.91 % single_blank_verbs
\\ 
& \small distilroberta 
& \small 1.74 % blank_verbs
& \small 2.18 % single_verb_antonyms
& \small 0.96 % single_external_verb_all
& \small 1.68 % single_blank_verbs
\\ 
& \small mpnet-base
& \small 1.85 % blank_verbs
& \small 2.37 % single_verb_antonyms
& \small \cellcolor{dustypink}{0.96} % single_external_verb_all
& \small 1.83 % single_blank_verbs
\\ \hline
\multirow{3}{0.4em}{\begin{turn}{90}
\small multi-qa - \end{turn}}&
\small distilbert 
& \small 1.91 % blank_verbs
& \small 2.26 % single_verb_antonyms
& \small 1.54 % single_external_verb_all
& \small 1.99 % single_blank_verbs
\\
& \small mpnet-base 
& \small 2.13 % blank_verbs
& \small 4.09 % single_verb_antonyms
& \small 1.3 % single_external_verb_all
& \small 2.17 % single_blank_verbs
\\ 
& \small MiniLM-L6 
& \small 1.63 % blank_verbs
& \small 1.58 % single_verb_antonyms
& \small 1.6 % single_external_verb_all
& \small 1.52 % single_blank_verbs
\\
\hline
\multirow{1}{0.4em}{\begin{turn}{90}
\small  \end{turn}}
& \small T5-gtr-base 
& \small 1.56 % blank_verbs
& \small 1.43 % single_verb_antonyms
& \small \cellcolor{dustypink}{0.1} % single_external_verb_all
& \small 1.47 % single_blank_verbs
\\ 
& \small nq-distilbert
& \small 1.86 % blank_verbs
& \small 2.45 % single_verb_antonyms
& \small 4.49 % single_external_verb_all
& \small 1.86 % single_blank_verbs
\\
& \small nli-distilrob.
& \small 1.79 % blank_verbs
& \small 1.35 % single_verb_antonyms
& \small \cellcolor{dustypink}{0.8} % single_external_verb_all
& \small 1.69 % single_blank_verbs
\\ 
\hline
\end{tabular}
\end{table*}